\documentclass[pdflatex,sn-mathphys-num]{sn-jnl}

\usepackage{graphicx}%
\usepackage{multirow}%
\usepackage{amsmath,amssymb,amsfonts}%
\usepackage{amsthm}%
\usepackage{mathrsfs}%
\usepackage[title]{appendix}%
\usepackage{xcolor}%
\usepackage{textcomp}%
\usepackage{manyfoot}%
\usepackage{booktabs}%
\usepackage{algorithm}%
\usepackage{algorithmicx}%
\usepackage{algpseudocode}%
\usepackage{listings}%
\DeclareRobustCommand{\method}[1]{{\fontsize{8}{12}\selectfont \textbf{#1}}}
\usepackage{adjustbox} 

\usepackage{amsmath}

\usepackage{caption}
\usepackage{subcaption}
\usepackage{makecell}

\usepackage{rotating}
\usepackage{adjustbox}
\usepackage{multirow}
\usepackage{tabularray}

\DeclareRobustCommand{\method}[1]{{\fontsize{8}{12}\selectfont \textbf{#1}}}

\usepackage{amsthm}
\usepackage{tabularx}

\theoremstyle{thmstyleone}%
\newtheorem{theorem}{Theorem}
\theoremstyle{thmstyletwo}%

\theoremstyle{thmstylethree}%

\begin{document}

\title[Article Title]{OLR-W\kern-0.25em A\kern-0.25em A: Adaptive and Drift-Resilient Online Regression with Dynamic Weighted Averaging}

\author*[1]{\fnm{Mohammad} \sur{Abu-Shaira}}\email{shairaabu-shaira@my.unt.edu}

\author[1]{\fnm{Weishi} \sur{Shi}}\email{weishi.shi@unt.edu}

\affil[1]{\orgdiv{Department of Computer Science and Engineering}, 
\orgname{University of North Texas}, 
\orgaddress{\city{Denton}, \state{TX}, \country{USA}}}









\abstract{Real-world datasets frequently exhibit evolving data distributions, reflecting temporal variations and underlying shifts. Overlooking this phenomenon, known as concept drift, can substantially degrade the predictive performance of the model. Furthermore, the presence of hyperparameters in online models exacerbates this issue, as these parameters are typically fixed and lack the flexibility to dynamically adjust to evolving data. This paper introduces ``OLR-W\kern-0.25em A\kern-0.25em A: An Adaptive and Drift-Resilient Online Regression with Dynamic Weighted Average", a hyperparameter-free model designed to tackle the challenges of non-stationary data streams and enable effective, continuous adaptation. The objective is to strike a balance between model stability and adaptability. OLR-W\kern-0.25em A\kern-0.25em A incrementally updates its base model by integrating incoming data streams, utilizing an exponentially weighted moving average. It further introduces a unique optimization mechanism that dynamically detects concept drift, quantifies its magnitude, and adjusts the model based on real-time data characteristics. Rigorous evaluations show that it matches batch regression performance in static settings and consistently outperforms or rivals state-of-the-art online models, confirming its effectiveness. Concept drift datasets reveal a performance gap that OLR-W\kern-0.25em A\kern-0.25em A effectively bridges, setting it apart from other online models. In addition, the model effectively handles confidence-based scenarios through a conservative update strategy that prioritizes stable, high-confidence data points. Notably, OLR-W\kern-0.25em A\kern-0.25em A converges rapidly, consistently yielding higher \text{R}$^2$ values compared to other online models. 
}





\maketitle

\section{Introduction}\label{sec1}

Traditional machine learning methods operate predominantly in a batch processing paradigm, where the complete dataset is assumed to be available for all computations. This framework enables iterative analysis, including multiple interactions with the data and rigorous cross-validation through subset exclusion. Moreover, it benefits from the absence of temporal constraints, allowing extensive training time, and assumes a stationary data distribution, typically modeled as independent and identically distributed (\textit{i.i.d.}). However, these assumptions \cite{fontenla2013online} present an oversimplified and often unrealistic representation of real-world data, making batch-based approaches less suitable for dynamic and evolving datasets.

The assumptions underlying batch learning significantly constrain its practical applicability, particularly in dynamic environments. For instance, a machine learning model trained to predict stock prices relies on historical market data, yet the inherent nonstationarity of financial markets causes predictive accuracy to degrade over time \cite{box2015time}. Economic fluctuations such as recessions and inflation, along with structural market changes such as new companies go public and existing ones file for bankruptcy, continuously reshape stock price behaviors. As a result, the initial training data become increasingly outdated, rendering the model less representative of current market conditions. Without regular updates and adaptation, its predictions lose reliability, highlighting the limitations of batch learning in evolving data environments

Modern streaming environments, characterized by continuous data arrival and the demand for timely predictions, render the batch paradigm impractical. Users cannot afford to wait for complete model retraining with each new data increment, necessitating a shift toward online learning. Unlike batch processing, online learning operates with only a subset of the data at any given time, provides timely responses, and adapts to evolving data distributions \cite{celik2023online}.  Consider, for instance, a smart city traffic management system.  Real-time predictions of traffic flow and congestion are crucial. Given the rapid fluctuations in traffic conditions due to accidents, road closures, or unforeseen events, continuous model updates with real-time traffic data, potentially exhibiting both temporal and spatial correlations, are essential.  This enables the system to provide drivers with accurate, up-to-the-minute information, empowering them to make informed routing decisions and thereby mitigating congestion in real time. The adoption of online models brings forth a transformative paradigm in predictive modeling, offering agility, efficiency, and adaptability in the face of ever-changing data landscapes \cite{shalev2012online}.

However, online learning encounter several challenges, including effectively managing \textit{concept drift} \cite{webb2016characterizing, ditzler2012incremental}, a prevalent issue in real-world streaming applications where data characteristics evolve unpredictably over time. These shifts can arise due to changing user behaviors, evolving market trends, seasonal variations, or unforeseen external factors, requiring models to continuously adapt while maintaining stability and predictive accuracy. Additionally, the presence of hyperparameters in online learning models presents a significant challenge, more specifically, with the existence of concept drift. Consequently, expecting users to manually recalibrate these parameters each time concept drift occurs is impractical \cite{barbaro2018tuning}.

Among online learning approaches, \textit{Linear Regression} and its online counterpart play a crucial role in data analysis and predictive modeling \cite{montgomery2021introduction}. Linear regression models the relationship between a dependent variable \( y \) and independent variables \( X \) as \( y = X\beta + \epsilon \), where \( \beta \) represents the regression coefficients and \( \epsilon \) is the error term, assumed to follow a normal distribution. Its online counterpart incrementally updates the model as new data arrives, making it well-suited for streaming environments. This adaptability is essential in fields such as finance for stock price forecasting, healthcare for predicting patient outcomes, marketing for demand forecasting, and manufacturing for quality control and process optimization.

Existing online regression methods face several limitations that hinder their effectiveness in dynamic, real-world environments. Many rely on fixed hyperparameters (such as learning rate or regularization), making them highly sensitive and ill-suited for non-stationary data with concept drift, since manually adjusting these parameters at each occurrence of drift is neither feasible nor practical. In addition to this lack of adaptivity, models like OSGD and OMGD, although classified as online methods due to their ability to process one data point or mini-batch at a time, retain previously seen data. This leads to increasing memory usage over time and contradicts the core principle of memory efficiency expected in true online learning. Another common limitation is the lack of a controlled decaying or weighting mechanism, which is essential for maintaining the balance between adaptability and stability. This is a fundamental objective in online learning and is particularly more important in confidence-based scenarios \cite{crammer2008exact, prasad2008decision}, where preserving trust in expert-verified points requires assigning them higher weights to ensure model stability. Moreover, most models do not include built-in drift detection and adaptation mechanisms, leaving them unprepared to respond effectively to evolving data distributions. 

This paper introduces OLR-W\kern-0.25em A\kern-0.25em A, a novel and versatile online linear regression model specifically designed to handle dynamic data streams in real-time environments. Key contributions include: (1) a novel strategy for dynamically redefining the model in response to emerging data patterns; (2) a proactive, in-memory drift detection and adaptation mechanism that employs variable thresholding and leverages key performance indicators (KPIs) without assuming any specific data distribution, making it suitable for high-dimensional and large-scale data streams; (3) the use of an Exponentially Weighted Moving Average with an automatically inferred decay factor, enabling adaptive response to drift while maintaining a balance between stability and adaptability; and (4) competitive performance on par with batch models in stationary regression scenarios, while effectively managing concept drift in dynamic environments, demonstrating the robustness and adaptability of OLR-W\kern-0.25em A\kern-0.25em A in real-time learning.

In conclusion, OLR-W\kern-0.25em A (Online Linear Regression with Weighted Average) is a novel online regression model designed for adaptive learning in streaming environments. It incrementally updates its base model using an exponentially weighted moving average to ensure smooth adaptation. The enhanced variant, OLR-W\kern-0.25em A\kern-0.25em A, where the final `A' stands for Adaptive, extends this approach by featuring a novel mechanism for real-time drift detection and automatic hyperparameter inference, enabling robust performance under evolving data distributions. Extensive evaluations across diverse benchmarks demonstrate that our method matches batch regression in static settings and consistently outperforms leading online models in dynamic scenarios. Moreover, OLR-W\kern-0.25em A\kern-0.25em A effectively bridges performance gaps caused by concept drift, further distinguishing it from existing approaches. The source code and datasets are publicly available at our GitHub repository \cite{OLR-WA_GIT}.

\section{Related Work}
Linear regression \cite{abu-mustafa2012, maulud2020review} is one of the most widely used and fundamental algorithms in statistical and machine learning. It models the relationship between one or more independent variables and a dependent variable, serving as a mathematical framework for predictive analysis. Additionally, in certain cases, it can be utilized to infer causal relationships. This section first provides a brief overview of batch regression before delving into a detailed exploration of online linear regression and its various models.  

Regression can be categorized into simple and multiple regression. Simple linear regression examines the relationship between two continuous variables, where \( x \) serves as the predictor (independent variable) and \( y \) as the response (dependent variable) \cite{maulud2020review}. It is mathematically represented as \( y = w_{0} + w_{1}x + \epsilon \), where \( w_0 \) and \( w_1 \) are model parameters and \( \epsilon \) represents the error term.  

Multivariate linear regression (MLR) extends this concept to multiple predictor variables, modeling the response as \( y = w_0 + w_{1}x_{1} + \cdots + w_{m}x_{m} + \epsilon \), where \( w_i \) (for \( i = 0, 1, ..., m \)) denote the coefficients associated with each predictor. In this formulation, \( y \) represents the observed dependent variable, while \( w_0 + w_{1}x_{1} + \cdots + w_{m}x_{m} \) corresponds to the predicted value, and \( \epsilon \) captures the residual error. The coefficients are typically estimated using the Moore-Penrose pseudoinverse, a widely used method for computing the least-squares solution \cite{maulud2020review}, as formalized in Equation \ref{eq:pseudo-inverse-main}, where \( \hat{W} \) denotes the estimated parameter values and \( N \) represents the number of data samples.

\begin{equation}
\hat{W} = (\textbf{X}^T\textbf{X})^{-1}\textbf{X}^T\mathbf{y}    
\label{eq:pseudo-inverse-main}
\end{equation}

\noindent
where 
$\textbf{W}$= 
$\begin{bmatrix}
w_{0} \\
w_{1} \\
\vdots\\
w_{M}
\end{bmatrix}$, \textbf{X}= $\begin{bmatrix}
    1 & x_{11} & x_{12} & \ldots & x_{1M} \\
    1 & x_{21} & x_{22} & \ldots & x_{2M} \\
    \vdots & \vdots & \vdots & \vdots & \vdots\\
    1 & x_{n1} & x_{n2} & \ldots & x_{NM} 
\end{bmatrix}$, \textbf{y} = $\begin{bmatrix}
    y_1\\
    y_2\\
    \vdots\\
    y_N    
\end{bmatrix}$

\vspace{10pt}

The Mean Squared Error (MSE) objective in linear regression is convex due to its quadratic form, ensuring a unique global minimum. This allows the problem to be efficiently solved using optimization techniques like gradient descent, improving both stability and accuracy \cite{boyd2004convex}. A more general principle for parameter estimation is Maximum Likelihood Estimation (MLE), a method that estimates the parameters of a statistical model by fitting a probability distribution to the data in a way that maximizes the likelihood function \cite{hastie2009elements}. The least squares approach used to determine model parameters in linear regression is a specific case of MLE \cite{bishop2006pattern}.

Beyond the regression methods discussed earlier, several alternative approaches cater to varying data characteristics and modeling requirements. Among these, Support Vector Regression (SVR) \cite{awad2015support} has proven to be a powerful technique for real-valued function estimation, designed to handle noise and uncertainty through Vapnik’s $\epsilon$-insensitive loss. It introduces a margin of tolerance where small errors are ignored, while larger deviations are penalized, enabling balanced treatment of over- and underestimations. SVR’s regularized objective promotes generalization by controlling model complexity and penalizing deviations beyond the margin \cite{smola2004tutorial}, making it effective across diverse, noisy datasets.

Statistical learning theory, online convex optimization (OCO), and game theory form the theoretical backbone of online learning techniques \cite{hoi2021online}. Convexity plays a central role in designing efficient algorithms for streaming data, as many online learning problems can be framed as OCO tasks. In this setting, an algorithm selects a decision from a convex set at each time step and incurs a loss defined by a convex function. The performance is evaluated through cumulative regret, which measures the gap between the algorithm's total loss and that of the best fixed decision in hindsight. Minimizing this regret is fundamental to ensuring the algorithm remains competitive over time.

In general, \textit{Online Learning} models adhere to the framework outlined in Algorithm~\ref{alg:generalonlineframework}, which is designed to iteratively update model parameters while minimizing the cumulative loss incurred over time.

\begin{algorithm}
\caption{General Online Learning Framework \cite{Alex2008}}
\label{alg:generalonlineframework}
\begin{algorithmic}[1]
\State \textbf{Initialize:} Set model parameters \( w_1 = 0 \)
\For{each round \( t = 1, \dots, T \)}
    \State Receive training instance \( x_t \)
    \State Predict label \( \hat{y}_t \in \mathbb{R} \)
    \State Observe true label \( y_t \in \mathbb{R} \)
    \State Compute loss \( l(\hat{y}_t, y_t, x_t) \)
    \State Update model parameters \( w_t \)
\EndFor
\end{algorithmic}
\end{algorithm}

This section provides a comprehensive analysis of prominent online linear regression algorithms, examining their strengths, limitations, and applicability. Through a rigorous comparison, we explore their underlying principles, performance characteristics, and real-world suitability, offering insights into advancements, challenges, and future research directions in the field.

\subsection{Online Linear Regression using Stochastic Gradient Descent (SGD)}

Stochastic Gradient Descent (SGD), or Online SGD \cite{ding2021efficient}, is a fundamental optimization method for large-scale and online learning \cite{bottou2010large}. It updates model parameters using a single data point per iteration, making it highly scalable with per-step complexity $O(M)$, where $M$ is the number of parameters \cite{jin2018regularizing}. SGD is commonly used to minimize Mean Squared Error (MSE) in regression tasks \cite{bishop2006pattern}, using the negative gradient direction for weight updates \cite{bottou1991stochastic}. Its stochastic nature introduces gradient noise that acts as implicit regularization, aiding generalization in streaming contexts \cite{hardt2016train}.

Despite its efficiency \cite{goodfellow2016deep, ruder2016overview}, SGD requires access to previously seen data, leading to memory overhead. It is sensitive to learning rate settings: large values can cause divergence, while small ones may slow convergence \cite{goodfellow2016deep, johnson2013accelerating}. SGD is also vulnerable to outliers \cite{shah2020choosing} and its implicit retention of past patterns makes it less suitable for non-stationary environments.

To improve convergence, adaptive methods like AdaGrad \cite{duchi2011adaptive}, RMSProp \cite{tieleman2012lecture}, and Adam \cite{kingma2014adam} adjust learning rates based on historical gradients. SGDM introduces a momentum term $\beta \in [0,1)$ to smooth updates and improve stability, especially in non-convex settings \cite{liu2020improved}. Additionally, mini-batch gradient descent offers a middle ground by reducing variance while maintaining scalability \cite{danner2015fully}.

\subsection{Online Regression using Mini-Batch Gradient Descent (MBGD)}

Mini-Batch Gradient Descent (MBGD) strikes a balance between Batch Gradient Descent (BGD) and Stochastic Gradient Descent (SGD), offering both computational efficiency and reduced gradient variance in online regression tasks \cite{dekel2012optimal, bottou2018optimization}. It updates weights using small batches of size $K$, allowing faster convergence and improved stability, especially in streaming and evolving data contexts. By tuning the batch size, practitioners control the trade-off between gradient noise and generalization \cite{li2014efficient, avrithis2021iterative}, with smaller batches increasing variance and larger ones reducing stochasticity \cite{montavon2012neural, ruder2016overview}. The overall time complexity is $O(I \cdot K \cdot M)$ for $I$ iterations and $M$ features, with $O(K \cdot M)$ per iteration \cite{j_nishchal_computational_2021}.

MBGD stabilizes learning in noisy, dynamic environments \cite{ruder2016overview} and benefits from adaptive optimizers like Adam and RMSprop, which adjust learning rates based on past gradients \cite{kingma2014adam, tieleman2012lecture}. Its inherent noise also acts as a regularizer, mitigating overfitting and promoting better generalization in online settings \cite{hoffer2017train}.

To scale MBGD for large and decentralized data, asynchronous and parallel variants like Hogwild! and Federated Learning have been introduced \cite{recht2011hogwild, mcmahan2017communication}. Momentum-based techniques and variance-reduction methods such as SVRG and SAGA further improve convergence and robustness under concept drift \cite{johnson2013accelerating, defazio2014saga}. Nonetheless, MBGD still requires careful tuning of fixed batch sizes and static optimizer settings under evolving data distributions.

\subsection{Widrow-Hoff (LMS)}

The Widrow-Hoff algorithm, also known as the Least Mean Squares (LMS) algorithm or Adaline, was introduced by Widrow and Hoff in 1960 \cite{widrow1960adaptive}. It is a foundational technique in online learning, adaptive filtering, and neural networks, particularly for linear regression and classification \cite{haykin2002adaptive}. LMS employs a stochastic gradient-based approach that iteratively updates weights using only the current data point, making it efficient for streaming settings \cite{fontenla2013online}.

LMS is computationally simple, avoiding matrix inversions required by more complex methods like Recursive Least Squares (RLS) \cite{hsia1983convergence}. With a per-iteration complexity of $O(P)$, where $P$ is the number of parameters, it is highly suitable for real-time applications \cite{fontenla2013online}.

To overcome LMS’s sensitivity to fixed learning rates, Normalized LMS (NLMS) introduces a time-varying step size proportional to the inverse energy of the input \cite{slock1993convergence, fontenla2013online}. More recent extensions integrate LMS with neural networks and reinforcement learning for nonlinear and autonomous system adaptation \cite{liu2011kernel}.

Despite its simplicity, LMS only minimizes only the current error without incorporating historical data, limiting its ability to capture long-term patterns \cite{fontenla2013online}. It also requires prior knowledge of input statistics for optimal learning rate selection, and may converge slowly compared to adaptive methods \cite{kwong1992variable}. NLMS can further introduce instability in highly dynamic environments due to potential overcorrection \cite{ghauri2013system}.

\subsection{Recursive Least Squares (RLS)}

Recursive Least Squares (RLS) is a classic adaptive filtering algorithm widely used in system identification and adaptive control \cite{fontenla2013online, islam2019recursive}. Unlike stochastic methods like LMS, RLS employs a deterministic update mechanism to minimize cumulative squared error over all seen data. Its use of the matrix inversion lemma allows rapid convergence, especially under correlated input signals, outperforming LMS in both accuracy and speed \cite{haykin2013adaptive}.

RLS’s key advantage lies in its fast adaptation and lack of a learning rate \cite{haykin2013adaptive, fontenla2013online}. The algorithm recursively updates a covariance matrix and parameter vector, ensuring convergence to the optimal solution equivalent to batch least squares \cite{bittanti1988deterministic}. A forgetting factor $\lambda$ controls how past data influences updates, enabling adaptability to nonstationary environments \cite{bittanti1988deterministic}. Its memory of all prior observations makes it effective in stable, low-noise conditions and scenarios with highly correlated inputs \cite{li2011lms}.

However, this performance comes at a cost, as RLS has quadratic time complexity $O(P^2)$, making it less practical for high-dimensional or resource-constrained environments \cite{haykin2013adaptive}. It is also prone to numerical instability due to recursive matrix operations, particularly in ill-conditioned or finite-precision settings \cite{douglas2000numerically}. Issues such as round-off errors and finite word-length effects can degrade reliability.

To mitigate these challenges, variants like Fast RLS (FRLS) \cite{xiao2007fast} and Sliding-Window RLS \cite{djigan2006multichannel} have been proposed, reducing complexity while maintaining accuracy. Additional stabilization methods such as regularization, QR decomposition, and square-root filtering have been developed to improve numerical robustness in practical applications \cite{liu1991dynamic}.

\subsection{Online Ridge Regression (ORR)}

Ridge regression introduces L2 regularization to mitigate overfitting and multicollinearity by penalizing large coefficients, thereby promoting model stability and reducing variance \cite{witten2013introduction}. Unlike ordinary least squares, it balances mean squared error minimization with coefficient shrinkage, making it effective for high-dimensional data \cite{hastie2009elements}. Online Ridge Regression (ORR) extends this principle to streaming settings, updating parameters iteratively and incorporating regularization directly into the stochastic gradient updates \cite{mohri2018foundations}.

ORR is well-suited for large-scale applications such as adaptive control, finance, and recommendation systems, offering linear time complexity per iteration \cite{j_nishchal_computational_2021}. It ensures real-time learning while maintaining regularization, which prevents overfitting and stabilizes updates over time.

However, ORR's effectiveness depends heavily on selecting an appropriate regularization parameter $\lambda$. A poorly chosen $\lambda$ can lead to underfitting or overfitting \cite{stephenson2021can}. To address this, adaptive techniques like Bayesian optimization \cite{snoek2012practical} and regularization path tracking \cite{huang2018unified} have been proposed to tune $\lambda$ dynamically.

Recent advancements have improved ORR's adaptability and robustness in nonstationary environments. Incremental kernel ridge regression with predictive sampling has shown near-optimal performance in online contexts \cite{xu2019new}. Other works introduced novel ridge estimators for high-dimensional regression under multicollinearity \cite{akhtar2024comparative}, and alternative algorithms such as forward-based stochastic methods have demonstrated better regularization resilience and theoretical guarantees \cite{ouhamma2021stochastic}.

\subsection{Online Lasso Regression (OLR)}

Lasso regression, or Least Absolute Shrinkage and Selection Operator, introduces L1 regularization to penalize large coefficients and enforce sparsity in linear models, enabling simultaneous regression and feature selection \cite{ranstam2018lasso, tibshirani1996regression}. The regularization parameter $\lambda$ controls the trade-off between accuracy and model simplicity \cite{hastie2009elements}.

Online Lasso Regression (OLR) extends this framework to streaming data by incrementally updating model parameters with each new observation \cite{mohri2018foundations}. It is particularly suitable for real-time applications, combining the adaptability of online learning with Lasso’s inherent feature selection, which efficiently manages complexity in high-dimensional environments.

OLR relies on stochastic gradient descent, which enables scalability for large and streaming datasets. However, its performance is sensitive to the regularization parameter. Poor tuning of $\lambda$ may lead to underfitting or overfitting \cite{hastie2009elements}. To address this challenge, adaptive methods such as data-driven regularization tuning with theoretical guarantees have been proposed \cite{chichignoud2016practical}.

Recent innovations include the Online Linearized Lasso by Yang et al. \cite{yang2023online}, a memory-efficient approach that improves convergence rates by constraining the solution path. For cases where Lasso struggles, especially when predictors outnumber observations, the elastic net offers a more robust alternative by combining L1 and L2 regularization \cite{zou2005regularization}. These developments enhance OLR’s reliability across domains like real-time analytics, finance, and recommendation systems.

\subsection{Online Passive-Aggressive (PA)}

The Online Passive-Aggressive (PA) algorithm \cite{crammer2006online} is an efficient margin-based online learning method closely related to Support Vector Machines. It updates model parameters only when a prediction error exceeds a predefined threshold, remaining passive when the loss is zero and aggressive otherwise. This selective update mechanism enables fast and focused learning for real-time regression tasks.

PA offers low time complexity ($O(M)$), quick convergence, and robustness to noise by emphasizing examples near the decision boundary \cite{crammer2006online}. Its updates are driven by an $\epsilon$-insensitive hinge loss, which ignores small errors and focuses updates on significant deviations. The algorithm has three main variants (PA-I, PA-II, PA-III), differing in how they compute the update magnitude based on regularization and aggressiveness parameters, allowing flexibility in learning dynamics \cite{crammer2006online}.

However, PA's performance is sensitive to the aggressiveness parameter $C$. Small values may slow convergence, while large values can cause overfitting due to overly aggressive updates \cite{crammer2006online}. The fixed threshold also makes it less adaptable in dynamic or noisy environments, limiting its responsiveness to evolving data patterns.

Recent enhancements address these limitations. Adaptive PA variants incorporate side information and dynamic threshold selection to improve convergence and robustness in non-stationary environments \cite{shi2024adaptive}. Other advancements apply adaptive regularization techniques to stabilize learning in high-dimensional settings and improve generalization under drift \cite{wustabilizing}.

\vspace{1cm}
In summary, while significant progress has been made in online linear regression, key limitations remain. These can be summarized as follows:
\begin{enumerate}
\item Most online learning algorithms rely on hyperparameters that are impractical to tune manually, especially in the presence of concept drift.
\item Many approaches lack a decay or weighting mechanism, which is crucial for balancing adaptability to new data with stability over time.
\item Existing methods are not drift-aware; they do not detect, quantify, or respond to distributional changes, nor do they adaptively optimize hyperparameters or update strategies accordingly.
\end{enumerate}

These limitations highlight the need for a more flexible and adaptive solution. This motivates the development of OLR-W\kern-0.25em A\kern-0.25em A, a drift-aware online regression framework that dynamically adjusts hyperparameters and model behavior to maintain an ideal balance between adaptiveness and stability in evolving data streams.

\section{OLR-W\kern-0.25em A Method}

This section presents a comprehensive overview of the ``OLR-W\kern-0.25em A: Online Regression with Weighted Average in Multivariate Data Streams," highlighting its core principles and distinctive features that establish it as a robust and efficient approach for online regression. It begins by outlining the problem settings, defining the online regression framework and the challenges posed by evolving data streams. Then, it explores the first variant, OLR-W\kern-0.25em A, detailing its parameter update mechanism and foundational characteristics. Following this, the discussion extends to OLR-W\kern-0.25em A\kern-0.25em A, showcasing its advancements over OLR-W\kern-0.25em A through the development of a novel memory-based automatic hyperparameter optimization mechanism, designed to enhance the model's adaptability in the presence of concept drift.

\subsection{Problem Settings}\label{problem-settings}

Consider \( \mathcal{X} \subseteq \mathbb{R}^d \) as the input feature space and \( \mathcal{Y} \subseteq \mathbb{R} \) as the continuous output space. At each time step \( t \), a new data instance \( (\mathbf{x}_t, y_t) \) is received, where \( \mathbf{x}_t \in \mathcal{X} \) represents the feature vector, and \( y_t \in \mathcal{Y} \) is the corresponding real-valued target. The data stream is assumed to be potentially infinite, requiring models to process and learn from data sequentially with limited memory resources \cite{gama2014survey}.

The goal of the online regression model is to iteratively update the predictive function \( f_t: \mathcal{X} \to \mathcal{Y} \) in order to minimize the cumulative regression error over time. In typical online learning scenarios, data arrives incrementally, allowing the model to adapt continuously as new observations are introduced. However, real-world data streams often exhibit non-stationarity, where the underlying data distribution changes over time, which is known as \textit{concept drift}. 

Formally, let \( \{(\mathbf{x}_1, y_1), (\mathbf{x}_2, y_2), \dots\} \) be a sequence of data points generated from an unknown joint probability distribution \( p(\mathbf{x}, y) \). Concept drift occurs when this joint distribution shifts between time steps, which can be mathematically expressed as below, where \( t_n \) and \( t_{n+1} \) denote consecutive time points, reflecting the dynamic nature of the data. 
\[
\exists \mathbf{x}, y : p_{t_n}(\mathbf{x}, y) \neq p_{t_{n+1}}(\mathbf{x}, y)
\]

Concept drift can occur in various forms, such as sudden (abrupt), gradual, or incremental changes, each presenting distinct challenges to maintaining model performance \cite{webb2016characterizing, gama2014survey}. Effectively managing concept drift is essential for preserving the performance and reliability of online regression models. This work aims to develop an adaptive online regression framework designed to accurately detect and respond to dynamic changes in data distributions, ensuring sustained predictive accuracy in evolving environments.

\subsection{Online Learning Parameter Update}\label{olr-wa1}

The OLR-W\kern-0.25em A framework facilitates incremental learning from continuous data streams by maintaining two distinct hyperplanes: the base hyperplane, \( f_{\text{base}}(x) \), representing the cumulative knowledge acquired from historical data, and the incremental hyperplane, \( f_{\text{inc}}(x) \), capturing insights derived from the current mini-batch at time \( t \), with \( x \) denoting the feature vector. The associated norm vectors, \( V_{\text{base}} \) and \( V_{\text{inc}} \), are combined through the Exponentially Weighted Moving Average (EWMA) method. A dynamic smoothing factor \( \alpha \in (0,1] \) regulates this combination, assigning a weight of \( \alpha \) to \( V_{\text{inc}} \) to prioritize recent information, while \( (1 - \alpha) \) preserves the influence of prior knowledge encapsulated in the base model. The aggregated norm vector, \( V_{\text{Avg}} \), undergoes continuous updates, utilizing the intersection point \( P_{\text{int}} \) of the base and incremental hyperplanes to refine the base model \( W_{\text{base}} \). This iterative process aims to incrementally align \( W_{\text{base}} \) with \( W_{\text{inc}} \) as more data becomes available, thereby enhancing the model’s adaptability and learning efficiency. The learning objective of our model is defined as follows:
\begin{align}
\mathbf{w}_{t+1} = \arg\min_{\mathbf{w}} \Bigg\{ 
    \frac{1}{2} \| \mathbf{w} - \mathbf{w}_t \|^2 + \alpha \mathcal{L}(y_{\text{inc}}, f(\mathbf{w}; \mathbf{X}_{\text{inc}}))
    \notag \\
    + (1 - \alpha) \mathcal{L}(y_{\text{base}}, f(\mathbf{w}; \mathbf{X}_{\text{base}}))
\Bigg\}
\end{align}

The objective function effectively balances \textit{stability} and \textit{adaptability} in an online regression setting. The first term, \( \frac{1}{2} \| \mathbf{w} - \mathbf{w}_t \|^2 \), acts as a regularization component, promoting smooth transitions between model updates by penalizing large deviations from the previous weight vector \( \mathbf{w}_t \). The second term, \( \alpha \mathcal{L}(y_{\text{inc}}, f(\mathbf{w}; \mathbf{X}_{\text{inc}})) \), represents the mean squared error (MSE) loss calculated over the incremental mini-batch. Here, \( \alpha \in (0,1] \) serves as a weighting factor that emphasizes recent data, thereby enhancing the model’s responsiveness to new patterns or shifts in the data distribution. Conversely, the term \( (1 - \alpha) \mathcal{L}(y_{\text{base}}, f(\mathbf{w}; \mathbf{X}_{\text{base}})) \) corresponds to the MSE loss associated with the base (historical) data. This term ensures that the model retains knowledge from previously encountered data, with \( (1 - \alpha) \) controlling the degree of emphasis on past observations. Together, these components enable the model to adapt efficiently to new information while maintaining robustness against concept drift.

OLR-W\kern-0.25em A utilizes the Exponentially Weighted Moving Average (EWMA) \cite{shumway2000time} to integrate the base and incremental models. At the start of each iteration, the weighting factor \((\alpha)\) is either user-defined or set to its default value \((\alpha = 0.5)\). In OLR-W\kern-0.25em A\kern-0.25em A, however, \((\alpha)\) is dynamically adjusted to optimize adaptability and robustness in response to concept drift, rendering OLR-W\kern-0.25em A\kern-0.25em A a hyperparameter-free model. This adaptive weighting strategy is elaborated in subsection \ref{olr-wa2}. The weighted average formula in Equation \ref{eq:olr-wa-gen-equation} maintains consistency across dimensions, integrating the norm vectors of the base and incremental hyperplanes using the dynamically computed \((\alpha)\) and its complement \((1 - \alpha)\):

\begin{equation}
\label{eq:olr-wa-gen-equation}
    V_{\text{Avg}} = \alpha \cdot V_{\text{inc}} + (1 - \alpha) \cdot V_{\text{base}} 
\end{equation}

In the realm of N-dimensional geometry, the definition of a new hyper-plane necessitates the utilization of a norm vector and a point \cite{stewart2020calculus}. Let \(\mathbf{n} = \langle n_1, n_2, \ldots, n_N \rangle\) denote the norm vector, representing the directional characteristics of the desired hyper-plane, and let \(\mathbf{P} = (X_{1_p}, X_{2_p}, \ldots, X_{N_p})\) be a point lying within the hyper-plane. To establish the equation of the new hyper-plane, consider an arbitrary point \(\mathbf{Q} = (x_1, x_2, \ldots, x_N)\) on the hyper-plane. The vector connecting \(\mathbf{P}\) and \(\mathbf{Q}\), denoted as \(\overrightarrow{\text{PQ}} = \langle x_1 - X_{1_p}, x_2 - X_{2_p}, \ldots, x_N - X_{N_p} \rangle\), lies within the hyper-plane and is orthogonal to the norm vector \(\mathbf{n}\). Hence, the orthogonality condition can be expressed as \(\mathbf{n} \cdot \overrightarrow{\text{PQ}} = n_1(x_1 - X_{1_p}) + n_2(x_2 - X_{2_p}) + \ldots + n_N(x_N - X_{N_p}) = 0\). This methodology enables the precise definition of a new hyper-plane in N-dimensional space based on its norm vector and a known point lying within the hyper-plane.

\begin{figure}[t]
\centering
\adjustbox{max width=\textwidth}{\includegraphics{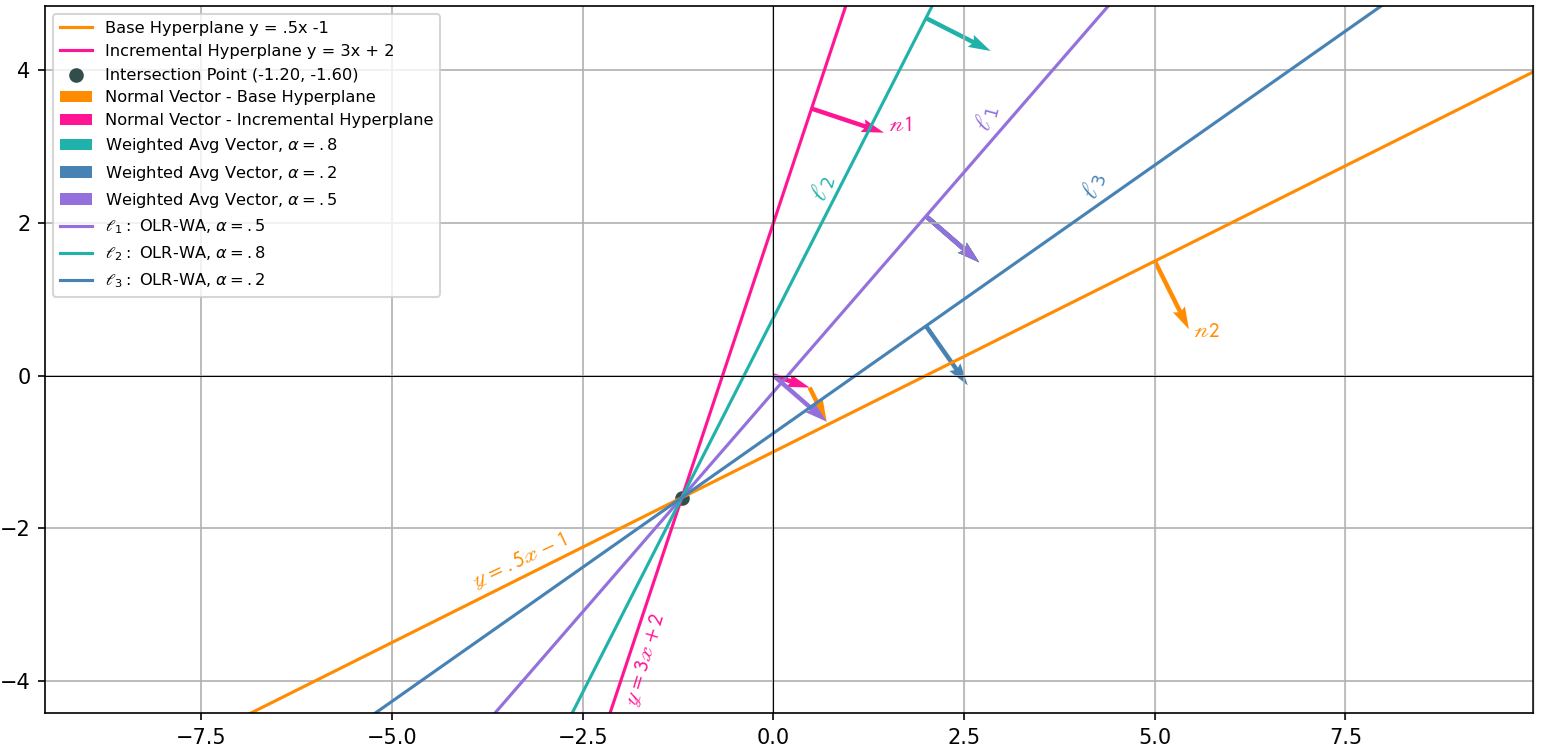}}
\caption{OLR-W\kern-0.25em A One Step Online Learning}
\label{fig:one_step_update}
\end{figure}

Figure \ref{fig:one_step_update} illustrates the process of defining the updated hyperplane within a two-dimensional space, which is then adopted as the new base model for the next iteration. This example reflects an adversarial scenario where the incoming data significantly diverges from the existing base model. The base hyperplane, shown in dark orange, represents the knowledge accumulated up to time \( (t - 1) \), while the current incoming batch at time \( (t) \) is highlighted in deep pink. The corresponding norm vectors for both hyperplanes are computed and visually represented using matching color schemes.

Assuming \( \alpha = 0.5 \), the weighted average vector, depicted in purple, is calculated using Equation \ref{eq:olr-wa-gen-equation}. The new hyperplane, labeled as line \( l_1 \), is established based on the intersection of the base and incremental boundaries, guided by the weighted average norm vector. Additionally, the figure highlights two more hyperplanes: \( l_2 \), corresponding to a higher \( \alpha \) value, which accelerates adaptation by giving more importance to new data, and \( l_3 \), representing a confidence-based strategy that prioritizes stability by focusing on data points with higher reliability. These boundaries showcase OLR-W\kern-0.25em A's flexibility in addressing diverse scenarios, including adversarial environments where a higher \( \alpha \) promotes rapid adaptation to significantly divergent data, and confidence-driven contexts where conservative updates ensure model stability. A practical example of the confidence-based approach is sentiment analysis, where data verified by domain experts or trusted sources can be assigned greater weights, reflecting their higher confidence and relevance.

The process of obtaining the point of intersection varies depending on the dimensionality of the space. For instance, in a two-dimensional setting, the intersection of two lines is found by solving a system of two equations with two variables, resulting in a single exact point of intersection. However, in three dimensions, determining the intersection of two planes involves solving a system of two equations with three variables, leading to an infinite number of solutions that represent the points forming the intersecting line. In higher dimensions, the intersection of N-Dimensional hyper-planes, is a (N-1)-dimensional hyper-plane which will be generated by solving two equations of N variables. This can be done using methods like Gaussian elimination, or using software packages. The solution will be a point $(x_1, x_2, ..., x_n)$ that satisfies both equations, representing the intersection of the two hyper-planes. Note that there can be multiple solutions, depending on the hyper-planes' orientation and position in the N-dimensional space. However, any solution is a accurate, to simply demonstrate this in a 2-dimensional space, a directional vector along with any point in the line of the intersection suffices to define the new hyperplane.

\begin{figure}[H]
    \centering    
    \begin{subfigure}[t]{0.49\textwidth}
        \centering
        \includegraphics[width=\textwidth]{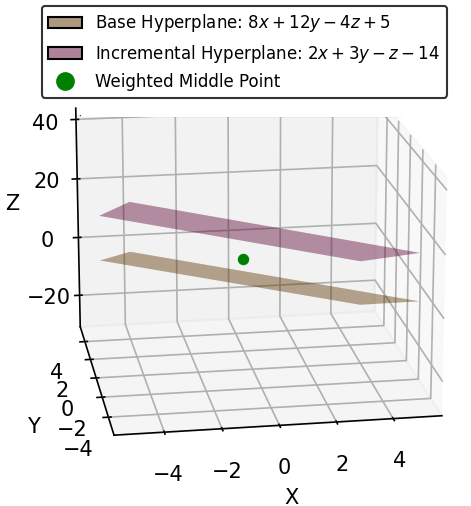}
        \caption{$\alpha=0.5$}
        \label{fig:parallel3}
    \end{subfigure}
    \hfill
    \begin{subfigure}[t]{0.49\textwidth}
        \centering
        \includegraphics[width=\textwidth]{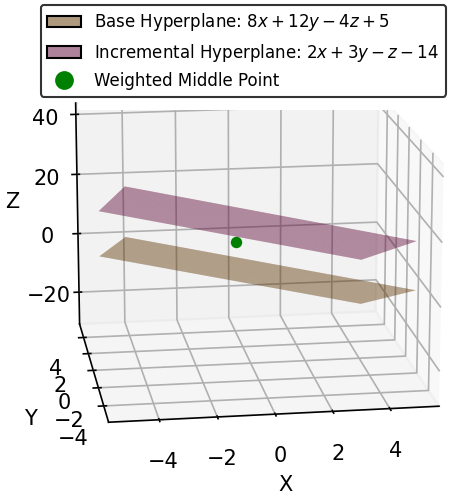}
        \caption{$\alpha=0.2$}
        \label{fig:parallel4}
    \end{subfigure}
    \caption{Weighted Middle Point}
    \label{fig:weighted-middle-point}
\end{figure}

The system relies on the presence of an intersection point. In the absence of one, two scenarios may occur, as described by Layton et al. \cite{layton2014numerical}. First, in the ``Coincide" case, no update is required as the base and incremental models are identical. Second, in the ``Parallel" case, the system calculates a weighted midpoint and uses it as the intersection point. Figure \ref{fig:weighted-middle-point} illustrates the concept of the weighted midpoint. In (a), with \(\alpha = 0.5\), the weighted midpoint is positioned exactly at the center. In (b), when \(\alpha = 0.8\), the weighted midpoint shifts towards the incremental model, along with the corresponding norm vector, collectively defining the new hyperplane.

\begin{algorithm}
    \caption{OLR-W\kern-0.25em A}
    \label{alg:olrwaalg}
    \begin{algorithmic}[1]

\State \textbf{Input:} Data stream \(\{ \mathbf{X}, \mathbf{y} \}\), \( \alpha = (0,1] \) , \quad \text{(default} \(\alpha\) = 0.5\text{)}

\State $\mathbf{W}_{\text{base}} \gets \arg\min\limits_{\mathbf{w}} \mathcal{L}_{\text{MSE}}(\mathbf{X}_{\text{base}}, \mathbf{y}_{\text{base}})$
\makebox[0pt][l]{\hspace{1cm}\Comment{\scriptsize \parbox{7cm}{(default 
 $\mathbf{X}_{\text{base}}^{\dagger} \mathbf{y}_{\text{base}}$),\\  $\mathbf{X}_{\text{base}}, \mathbf{y}_{\text{base}}$ is the first mini-batch}}}

\For{$t \gets 1$ to $T$}

\State $\mathbf{W}_{\text{inc}} \gets \arg\min\limits_{\mathbf{w}} \mathcal{L}_{\text{MSE}}(\mathbf{X}_{\text{inc}}, \mathbf{y}_{\text{inc}})$

\If{ $\frac{\mathbf{W}_{\text{base},i}}{\mathbf{W}_{\text{inc},i}} = \text{constant} \quad \forall i $ }\makebox[0pt][l]{\hspace{1.2cm}\Comment{\scriptsize \parbox{7cm}{Check if the hyperplanes coincide}}}
    \State \textbf{continue}
\EndIf

\State $\mathbf{P}_{\text{int}} \gets$ \method{ FindIntersectPoint} $(\mathbf{W}_{\text{base}}, \mathbf{W}_{\text{inc}})$

\State $\mathbf{V}_{\text{base}} \gets \method{ComputeNormVector}(\mathbf{W}_{\text{base}})$
\State $\mathbf{V}_{\text{inc}} \gets \method{ComputeNormVector}(\mathbf{W}_{\text{inc}})$

\State $\mathbf{V}_{\text{base}} \gets \frac{\mathbf{V}_{\text{base}}}{\|\mathbf{V}_{\text{base}}\|_2}$
\quad\quad
$\mathbf{V}_{\text{inc}} \gets \frac{\mathbf{V}_{\text{inc}}}{\|\mathbf{V}_{\text{inc}}\|_2}$

\State $\mathbf{V}_{\text{avg}} \gets (1 - \alpha) \mathbf{V}_{\text{base}} + \alpha \mathbf{V}_{\text{inc}}$

    \State $\mathbf{W}_{\text{base}} \gets \method{DefineHyperplane}(\mathbf{V}_{\text{avg}}, \mathbf{P}_{\text{int}})$

\EndFor

\State \textbf{return} $\mathbf{W}_{\text{base}}$
\end{algorithmic}
\end{algorithm}
\subsection{Normalizing the Norm Vectors}
The norm vector encapsulates both magnitude and direction, with its magnitude varying depending on the coefficients of the corresponding hyperplane. Consequently, \textit{normalizing} the norm vectors is a critical step to ensure consistency and comparability. Consider the scenario where \( \alpha = 0.5 \), and the weighted average operation simplifies to a summation, represented by the placement of norm vectors at the origin. Without normalization, directly utilizing the norm vectors in their original form would yield a significantly different average norm vector, thereby substantially impacting the influence of the weighting factor \( \alpha \).

Furthermore, normalizing the vectors is a critical step in controlling potential overfitting, a phenomenon where the model fits the training data too closely, capturing noise and thereby reducing its ability to generalize to unseen data. By normalizing the norm vectors, OLR-W\kern-0.25em A effectively mitigates this risk, ensuring that the model prioritizes learning the underlying patterns rather than memorizing the training data.
\subsection{Memory-Based Automatic Hyperparameters Optimization} \label{olr-wa2}

The presence of hyperparameters in online learning presents significant challenges, primarily due to the evolving nature of data distributions over time, a phenomenon known as concept drift \cite{lu2018learning}. In such scenarios, it is impractical for users to continuously update hyperparameters with each occurrence of concept drift. Online learning models are inherently designed to adapt to dynamic data streams; however, their ability to do so varies considerably. Some models experience substantial performance degradation at the point of drift but rapidly adjust to the new concept, whereas others require a prolonged period to effectively accommodate the distributional shift and achieve adaptation.

OLR-W\kern-0.25em A\kern-0.25em A introduces an in-memory, built-in hyperparameter optimization technique designed for automatic drift detection and adaptation in dynamic data streams. This method leverages memory resources and key performance indicators (KPIs) without assuming any specific underlying data distribution, making it well-suited for managing high-dimensional datasets and efficiently processing large-scale data. The framework is integrated with the Self-Adjusting Memory (SAM) architecture, enhancing real-time processing capabilities and adaptability. Additionally, it incorporates variable thresholding techniques, further reinforcing its effectiveness in dynamically evolving environments.

OLR-W\kern-0.25em A\kern-0.25em A's hyperparameter optimization technique is designed to achieve four key objectives. First, it facilitates drift detection by identifying shifts in the underlying data streams. Second, it quantifies the magnitude of these drifts, offering valuable insights into their significance. Third, it employs a hyperparameter adjustment mechanism that leverages fine-tuned parameters to dynamically adapt the model to immediate fluctuations. Lastly, it ensures model calibration using the optimized hyperparameters to maintain performance consistency. The optimization process is detailed in Algorithm \ref{alg:olrwaalg2}, beginning at line 15.

OLR-W\kern-0.25em A\kern-0.25em A systematically captures Key Performance Indicators (KPIs) from incoming data streams, which can be processed as mini-batches or individual data points, denoted as (\(\mathbf{X}_{\text{inc}}, \mathbf{y}_{\text{inc}}\)). The KPIs for each processed mini-batch are stored in a compact, sequentially structured repository referred to as the KPI-Window (KPI-Win). These KPIs may include metrics such as Accuracy, Precision, Recall, and F1-score, among others. The selection of KPIs is flexible and depends on the specific problem domain. For instance, in \textit{regression} tasks, commonly used performance metrics include Mean Absolute Error (MAE), Mean Squared Error (MSE), Root Mean Squared Error (RMSE), and R-squared (\(R^2\)).

The size of the KPI-Window (\textit{KWS}) is a bounded variable, determined by the formulation presented in Equation \ref{eq:window-size}. In online learning settings, where the number of incoming data samples is theoretically infinite (\(\infty\)), the parameter \( N \) represents the number of the most recent data points retained within the rolling window, while \( K \) denotes the mini-batch size. The scaling factor (\(\delta \)) is selected based on specific application requirements, with an empirically chosen value of \( \delta = 0.05 \). The \textit{KWS} is constrained within predefined lower (LB) and upper (UB) bounds; any values falling below LB are set to LB, whereas values exceeding UB are capped at UB to ensure stability and efficient resource utilization.

The bounds LB and UB are introduced to prevent the KPI-Window from becoming either too narrow or too wide. A very narrow window with insufficient observations may result in noisy KPI estimates, reducing the reliability of performance monitoring. On the other hand, an excessively large window may include outdated observations, which can lead to inaccurate predictions at times when adaptability is crucial in online learning. Additionally, a large window may delay the model’s response to concept drift and increase computational overhead. These bounds are therefore essential to maintain a balance between stability and responsiveness. Empirically, we found that setting the lower bound to 11 and the upper bound to 31 achieves this balance effectively across various datasets with different drift conditions. These values were selected through grid-search validation to ensure both reliable adaptation and efficient memory use.

\begin{equation}
    \label{eq:window-size}
    \begin{aligned}
        &\text{LB} \leq \text{KWS}
        &\quad= \left(\frac{N}{K}\right) \times \delta \leq \text{UB}
    \end{aligned}
\end{equation}

\begin{algorithm}
    \caption{OLR-W\kern-0.25em A\kern-0.25em A}
    \label{alg:olrwaalg2}
    \begin{algorithmic}[1]

\State \textbf{Input:} Data stream \(\{ \mathbf{X}, \mathbf{y} \}\), \( \alpha = (0,1] \) \text{(default} \(\alpha\) = 0.5\text{)}, \( z \approx [1.5, 2.5] \), \\
\hspace{34pt}\( \textit{kpi} \subseteq \{\text{$R^2$, Loss, etc.}\} \)

\State $\mathbf{W}_{\text{base}} \gets \arg\min\limits_{\mathbf{w}} \mathcal{L}_{\text{MSE}}(\mathbf{X}_{\text{base}}, \mathbf{y}_{\text{base}})$
\makebox[0pt][l]{\hspace{1cm}\Comment{\scriptsize \parbox{7cm}{(default 
 $\mathbf{X}_{\text{base}}^{\dagger} \mathbf{y}_{\text{base}}$),\\  $\mathbf{X}_{\text{base}}, \mathbf{y}_{\text{base}}$ is the first mini-batch}}}

\For{$t \gets 1$ to $T$}

\State $\mathbf{W}_{\text{inc}} \gets \arg\min\limits_{\mathbf{w}} \mathcal{L}_{\text{MSE}}(\mathbf{X}_{\text{inc}}, \mathbf{y}_{\text{inc}})$

\If{ $\frac{\mathbf{W}_{\text{base},i}}{\mathbf{W}_{\text{inc},i}} = \text{constant} \quad \forall i $ }\makebox[0pt][l]{\hspace{1.2cm}\Comment{\scriptsize \parbox{7cm}{Check if the hyperplanes coincide}}}
    \State \textbf{continue}
\EndIf

\State $\mathbf{P}_{\text{int}} \gets$ \method{ FindIntersectPoint} $(\mathbf{W}_{\text{base}}, \mathbf{W}_{\text{inc}})$

\State $\mathbf{V}_{\text{base}} \gets \method{ComputeNormVector}(\mathbf{W}_{\text{base}})$
\State $\mathbf{V}_{\text{inc}} \gets \method{ComputeNormVector}(\mathbf{W}_{\text{inc}})$

\State $\mathbf{V}_{\text{base}} \gets \frac{\mathbf{V}_{\text{base}}}{\|\mathbf{V}_{\text{base}}\|_2}$
\quad\quad
$\mathbf{V}_{\text{inc}} \gets \frac{\mathbf{V}_{\text{inc}}}{\|\mathbf{V}_{\text{inc}}\|_2}$

\State $\mathbf{V}_{\text{avg}} \gets (1 - \alpha) \mathbf{V}_{\text{base}} + \alpha \mathbf{V}_{\text{inc}}$

    \State $\mathbf{W}_{\text{base}} \gets \method{DefineHyperplane}(\mathbf{V}_{\text{avg}}, \mathbf{P}_{\text{int}})$

    \State \scriptsize \texttt{/* Start hyperparameter optimization once the} \\
    \hspace{22pt} \texttt{KPI window reaches its threshold size. */}
    
    \State $\mathbf{MB}_{\text{kpi}} \gets \method{MbKPI}(\mathbf{X}_{\text{inc}}, \mathbf{y}_{\text{inc}}, \mathbf{W}_{\text{base}}, \textit{{kpi}})$

    \State \method{AddItem}($L$, $\mathbf{MB}_{\text{kpi}}$)

    \State $ \tau, \mu, \sigma, \text{low}, \text{high}, \text{DM} \gets \method{MeasureKPIs}(L, z)$

    \State $\textit{drift} \gets \method{DetectDrift}(\mu, L(-1), \tau)$ 

    \If{\textit{drift}}  
        \State \method{RemoveItem}($L$, $-1$)
        
        \State $\text{SM} \gets \method{DefineScaleMap}(\mu, \text{low}, \text{high})$ \Comment{SM is the scale map}
        \State $\alpha' \gets \method{TuneHyperparams}(\text{SM}, \text{DM})$

        \State $\mathbf{V}_{\text{avg}} \gets (1 - \alpha') \mathbf{V}_{\text{base}} + \alpha' \mathbf{V}_{\text{inc}}$

        \State $\mathbf{W}_{\text{base}} \gets \method{UpdateHyperplane}(\mathbf{V}_{\text{avg}}, \mathbf{P}_{\text{int}})$
    \EndIf
    
\EndFor

\State \textbf{return} $\mathbf{W}_{\text{base}}$
\end{algorithmic}
\end{algorithm}

\begin{figure}[ht]
  \centering
  \includegraphics[width=0.7\textwidth, height=0.3\textheight, keepaspectratio]{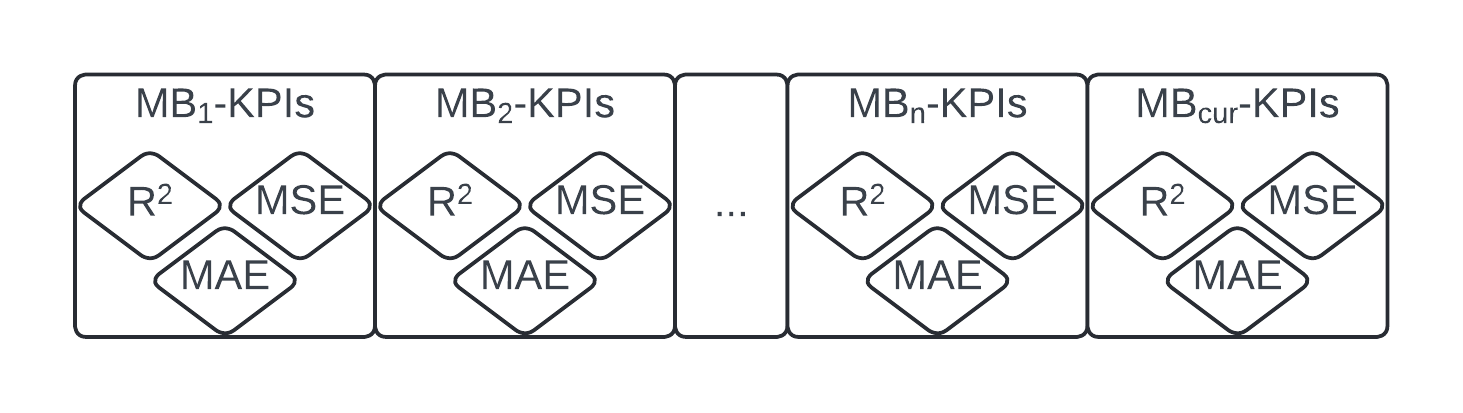}
  \caption{KPIs Window}
  \label{fig:002_KPIs_Window}
\end{figure}

In an illustrative scenario featuring 31 entries in KPI-Win, where each entry encompasses a set of KPIs related to the consumed mini-batch, the initial 30 entries are allocated for historical data. The final entry is designated for the current mini-batch, denoted as CUR-MB. The CUR-MB entry is transient and dynamically updates in response to concurrent computations, while the remaining entries represent our \textit{Baseline Statistics}, serving as a point of reference for comparing performance changes. Figure \ref{fig:002_KPIs_Window} illustrates the data structure utilized within OLR-W\kern-0.25em A\kern-0.25em A, where each entry represents a bag comprising one or more KPIs.

OLR-W\kern-0.25em A\kern-0.25em A structures historical KPIs within a normal distribution to facilitate the calculation of a dynamic threshold \( \tau \). This dynamic approach circumvents the limitations of fixed thresholds, which may not be universally applicable across different KPIs or data streams due to their diverse characteristics and evolving patterns. The threshold \( \tau \) is determined using Equation \ref{eq:threshold}, where \( \sigma \) represents the standard deviation of the distribution, and \( z \) is a scaling factor that typically corresponds to the number of standard deviations from the mean. The value of \( z \) is user-defined, with a commonly accepted range between 1.5 and 2.5. A lower \( z \) value, typically below 1.5, can be selected to identify subtle variations, and subsequently leading to the inference of a higher smoothing factor \(\alpha\), thereby ensuring a greater level of adaptation. Conversely, a higher \( z \) value adopts a more conservative approach, allowing for less sensitivity to minor fluctuations.

\begin{figure}[ht]
  \centering
  \includegraphics[width=0.7\textwidth, height=.35\textheight]{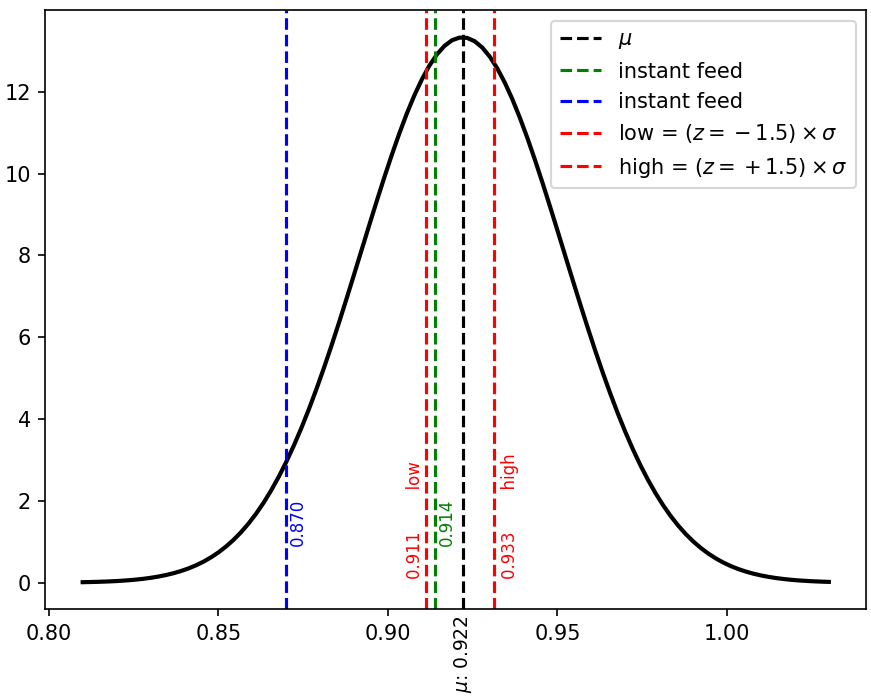}
  \caption{Limits - Gauss Dist. of KPI-Win on KPI = $R^2$}
  \label{fig:003_SCCM_LIMITS}
\end{figure}

\begin{figure}[ht]
  \centering
  \includegraphics[width=0.7\textwidth, height=.27\textheight, keepaspectratio]{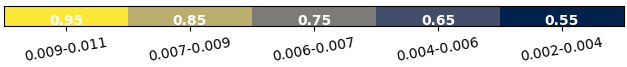}
  \caption{Scale Map}
  \label{fig:004_SCCM_SCALE}
\end{figure}

The proposed mechanism determines the optimal hyperparameter adjustments based on the detected drift magnitude, which is calculated using Equation \ref{eq:drift_magnitude_lstm_sccm}. In this context, \text{\Large $\mu$}$_{\scriptscriptstyle\text{KPI}}$ represents the mean of the KPI-Window (KPI-Win) for the specified KPI, while \text{CUR-MB}$_{\scriptscriptstyle\text{KPI}}$ denotes the KPI value of the current data feed under evaluation. Furthermore, the mechanism establishes two critical limits, referred to as `low' and `high,' which are derived from the computed variable threshold using Equations \ref{eq:low} and \ref{eq:high}, and as illustrated in Figure \ref{fig:003_SCCM_LIMITS}. Data feeds that fall within these thresholds, between the mean and one of the boundaries, are regarded as minor deviations, indicative of incremental drift. Conversely, data feeds that exceed these thresholds signal significant deviations, identifying abrupt drift. Figure \ref{fig:004_SCCM_SCALE} presents an illustrative example of a dynamically constructed scale map, which is updated iteratively. Given that the range of the smoothing factor \( \alpha \) is defined as \((0,1]\), and the drift magnitude is quantified, the appropriate value of \( \alpha \) can be inferred accordingly.

\begin{equation}
    \label{eq:threshold}
    \text{Threshold } (\tau) = z \times \sigma
\end{equation}
\begin{equation}
    \label{eq:low}
\text{low}  = \text{\Large $\mu$}_{\scriptscriptstyle\text{KPI}} - \tau  
\end{equation}
\begin{equation}
    \label{eq:high}
     \text{high} = \text{\Large $\mu$}_{\scriptscriptstyle\text{KPI}}+ \tau 
\end{equation}
\begin{equation}            
\label{eq:drift_magnitude_lstm_sccm}
    \text{DM} = \text{\Large $\mu$}_{\scriptscriptstyle\text{KPI}} - \text{ CUR-MB}_{\scriptscriptstyle\text{KPI}}
\end{equation}


In certain scenarios, the nature of the learning task can be anticipated in advance, allowing for the selection of an appropriate adaptation strategy. For instance, if the task is \textit{time-based}, the model is expected to continuously adapt to evolving patterns over time. Alternatively, if the task is \textit{confidence-based}, the existing model is already well-established and highly reliable, thereby carrying significant weight in decision-making. In such cases, employing a fixed smoothing factor, whether a high value (e.g., 1) or a low value (e.g., 0.1), may initially appear effective. A high smoothing factor places greater emphasis on recent data, enabling rapid adaptation to changes, whereas a low smoothing factor prioritizes historical knowledge, ensuring stability over time.

However, using a fixed smoothing factor presents two critical challenges. First, a high smoothing factor can result in \textit{overfitting}, where the model becomes excessively sensitive to recent changes, potentially capturing noise instead of meaningful trends. Furthermore, it increases the risk of \textit{ catastrophic forgetting}, where the model fails to retain valuable past knowledge, leading to suboptimal long-term performance. Second, a low smoothing factor will result in very low adaptability in responding to rapid changes in the data stream, as it places greater emphasis on historical data while minimizing the influence of recent observations. This can lead to delayed adjustments in dynamic environments, reducing the model's ability to effectively capture and react to sudden shifts or concept drifts. In response, the proposed OLR-W\kern-0.25em A\kern-0.25em A  balances \textit{adaptiveness} (responding to new changes) with \textit{robustness} (preserving learned knowledge), ensuring responsiveness to evolving data while maintaining stability.
\subsection{Theoretical Analysis of Convergence}
\subsubsection*{Model Description}
The OLR-W\kern-0.25em A model incrementally learns from streaming data by maintaining two hyperplanes: the base hyperplane \( f_{\text{base}}(x) \), which encapsulates the accumulated knowledge from past observations, and the incremental hyperplane, \( f_{\text{inc}}(x) \), which reflects the knowledge derived from the current mini-batch (instant feed) at time \( t \), where \( x \) denotes the feature vector. The corresponding norm vectors of these boundaries, denoted as \( V_{\text{base}} \) and \( V_{\text{inc}} \), respectively, are integrated using the Exponentially Weighted Moving Average (EWMA) technique. The dynamic weighting factor \( \alpha \in (0,1] \) serves as a smoothing parameter, assigning a weight of \( (\alpha) \) to the incremental norm vector, emphasizing the contribution of the most recent data, while the remaining weight \( (1 - \alpha) \) accounts for the historical knowledge retained in the base model. The combined norm vector \( V_{\text{Avg}} \) is iteratively updated, leveraging the intersection point \( P_{\text{int}} \) to redefine the base norm vector, with the objective of progressively aligning with \( V_{\text{inc}} \) as the number of observations increases.

\subsubsection*{Assumptions}
\begin{enumerate}
    \item \textbf{Linearity:} The relationship between the features and the target variable follows a linear model.    
    \item \textbf{Concept Stability} \cite{buzmakov2014concept}\textbf{:} The underlying data distribution is subject to temporal evolution and may experience various forms of drift. However, it is assumed that each incoming mini-batch represents a coherent and stable concept, meaning that the data within the mini-batch pertains to a single, consistent concept rather than being fragmented across multiple evolving concepts. This assumption ensures that the model can effectively adapt to gradual changes without ambiguity.
    \item \textbf{Boundedness of Data:} The input features and target variable are confined within a finite range, ensuring the model's stability and preventing unrealistic predictions beyond the defined input domain.
    \item \textbf{Continuity:} Hyperplanes are continuous functions.
\end{enumerate}

\subsubsection*{Convergence Criterion}
Our objective is to demonstrate the asymptotic convergence of the OLR-W\kern-0.25em A model, specifically focusing on the intersection point and the weighted average norm vector, as the number of observations approaches infinity.

\noindent
\textbf{Step 1: Convergence of Incremental Norm Vector \( V_{\text{inc}} \):}

Under the assumption of concept stability \cite{buzmakov2014concept}, the norm vector \( V_{\text{inc}} \) of the new incoming mini-batches (instant feeds) is expected to adapt and approximate optimal values over time as \( t \rightarrow \infty \), facilitating improved classification performance for any given instant feed at time \( t \).

\noindent
\textbf{Step 2: EWMA-Based Integration and Computation of the Weighted Average Norm Vector:}

Upon receiving the regression hyperplanes from both the base and incremental models at every time point \( t \), where the base model encapsulates the past accumulated knowledge from all previous observations and the incremental model reflects the knowledge derived from the current mini-batch, the OLR-W\kern-0.25em A framework computes their corresponding norm vectors, denoted as \( V_{\text{base}} \) and \( V_{\text{inc}} \). To effectively capture the evolving hyperplanes, OLR-W\kern-0.25em A combines these norm vectors to derive a weighted average norm vector, \( V_{\text{Avg}} \), at each iteration \( t \). This vector encapsulates both the direction and magnitude of the evolving classification model.

The integration process employs the Exponentially Weighted Moving Average (EWMA) approach, where the weighting factor \( \alpha \) determines the relative importance of the most recent incremental mini-batches (instant feeds), while \( 1 - \alpha \) accounts for previously accumulated knowledge. The weighting factor \( \alpha \) serves as an adaptation parameter, allowing the model to balance the responsiveness to new information with the retention of previous learning.

The weighted average norm vector, \( V_{\text{Avg}} \), is computed as follows:
\[
V_{\text{Avg}} = \alpha \cdot V_{\text{inc}} + (1 - \alpha) \cdot V_{\text{base}}
\]

\noindent
\textbf{Step 3: Convergence of OLR-W\kern-0.25em A Model}

\noindent
\textbf{Convergence of} \( V_{\text{Avg}} \) \textbf{to} \( V_{\text{inc}} \)

\noindent
Let \( \Delta_t = \| V_{\text{Avg}} - V_{\text{inc}} \| \) represent the norm of the difference between the weighted average norm vector and the incremental norm vector associated with the evolving hyperplanes. We aim to prove that \( \Delta_t \rightarrow 0^+ \) as \( t \rightarrow \infty \).

\begin{theorem}
Let \( t \) denote the number of iterations. Let \( \alpha \) denote the smoothing parameter. As \( t \rightarrow \infty \), the EWMA converges to the true mean or central tendency of the observed process.
\end{theorem}

\subsubsection*{Proof Sketch:}
\begin{enumerate}
    \item Recall that in the OLR-W\kern-0.25em A model, the weighted average norm vector \( V_{\text{Avg}} \) is obtained by combining the norm vectors of the base and incremental regression hyperplanes using Exponentially Weighted Moving Averages (EWMA) with dynamic weighting factor \( \alpha \).
    \item Therefore, \( V_{\text{Avg}} = \alpha \cdot V_{\text{inc}} + (1 - \alpha) \cdot V_{\text{base}} \).
    \item The weighted average norm vector is computed iteratively using the EWMA formula to ensure an adaptive balance between recent and historical information.
\end{enumerate}

By systematically incorporating information from all past observations, an iterative application of the EWMA formulation results in the following expansion:

\begin{equation}
 \bar{X}_t  = (1 - \alpha) \bar{X}_{t-1} + \alpha X_t   
\end{equation}

\[
\bar{X}_t = (1 - \alpha) \bar{X}_{t-1} + \alpha X_t
\]

\[
\bar{X}_t = (1 - \alpha)\left[(1 - \alpha) \bar{X}_{t-2} + \alpha X_{t-1}\right] + \alpha X_t
\]

\[
\bar{X}_t = (1 - \alpha)^2 \bar{X}_{t-2} + \alpha(1 - \alpha) X_{t-1} + \alpha X_t
\]

\[
\bar{X}_t = (1 - \alpha)^2\left[(1 - \alpha) \bar{X}_{t-3} + \alpha X_{t-2}\right] + \alpha(1 - \alpha) X_{t-1} + \alpha X_t
\]

\[
\bar{X}_t = (1 - \alpha)^3 \bar{X}_{t-3} + \alpha(1 - \alpha)^2 X_{t-2} + \alpha(1 - \alpha) X_{t-1} + \alpha X_t
\]

\[
\vdots
\]

\[
\bar{X}_t = (1-\alpha)^{t} \bar{X}_{0} + \alpha \sum_{k=0}^{t-1} (1-\alpha)^{k} X_{t-k}
\]


\[
V_{\text{Avg}} = (1-\alpha)^{t} V_{\text{base}} + \alpha \sum_{k=0}^{t-1}  (1-\alpha)^{k} V_{\text{inc}, t-k}
\]
Through exponential smoothing, the assigned weights decrease exponentially as observations recede further into the past. The smoothing parameter \( \alpha \) controls the rate of decay, with higher values favoring recent observations and lower values retaining historical influence over a longer period.

Applying the same principles to the OLR-W\kern-0.25em A model, it follows that as \( V_{\text{Avg}} \) approaches \( V_{\text{inc}} \), the hyperplane defined by the intersection point converges, ensuring that \( \Delta_t \rightarrow 0^+ \).
\subsection{Time and Memory Complexity}
\subsubsection*{Time Complexity}
OLR-W\kern-0.25em A utilizes Moore-Penrose pseudo-inverse once in the computation of the base hyperplane, and \((T)\) times for the computation of the incremental hyperplanes, using the Moore-Penrose pseudo-inverse is a particularly common choice as it is polynomial time. Given X to be a $N$ by $M$ matrix, where $N$ is the number of samples and $M$ the number of features. The matrix multiplications each require $O(NM^2)$,
while multiplying a matrix by a vector is $O(NM)$. Computing the inverse requires $O(M^3)$ in order to compute the LU or (Cholesky) factorization. 
Asymptotically, $O(NM^2)$ dominates $O(NM)$, so this calculation is ignored. Since the normal equation is used, the assumption that N $>$ M holds,  otherwise the matrix $\textbf{X}^T\textbf{X}$ would be singular (and hence non-invertible), which means that $O(NM^2)$ asymptotically dominates $O(M^3)$. Therefore, the total complexity for the pseudo inverse is $O(NM^2)$. 

OLR-W\kern-0.25em A processes data incrementally, enabling efficient computation by handling smaller data batches. Unlike the batch version, the online OLR-W\kern-0.25em A model achieves a lower time complexity, estimated at approximately \(O(KM^2)\) per iteration, where \(K\) represents the number of samples in the mini-batch. This incremental approach not only reduces computational overhead but also ensures faster and more scalable processing, particularly in real-time or large-scale data environments.

This analysis presents the complexity in a ``static context''; however, it is important to consider the practical implications. In practice, the batch model would require reprocessing all existing and new data points if encountering new data, resulting in a significantly higher time complexity compared to online models. This is primarily because the batch model's computational burden increases with the growing size of data points (N) when new data is introduced, while, in online, K remains fixed as a small mini-batch throughout.

OLR-W\kern-0.25em A method comprises several procedural stages, the computational complexities of which are disregarded, as it is dominated by the pseudo-inverse computation $O(KM^2)$. These stages are succinctly outlined as follows:
\begin{enumerate}
\item Hyper-plane Coincidence Check: The initial step involves confirming the alignment of two hyper-planes. This verification process, entailing iterations solely over specified norm vectors, demonstrates a time complexity of $O(M)$, where M signifies the dimensionality of the space.
\item Linear Equation System Solution: The resolution of a linear equation system can be achieved through various methodologies, such as Gaussian elimination. In the context of this method, which necessitates solving a system with two equations, the highest order time complexity is $O(N^3)$ \cite{jeannerod2013rank}. where N is the number of equations, since OLR-W\kern-0.25em A consists of only two equations, solving a system of two linear equations will be constant.
\item Weighted Midpoint Determination (Parallel Planes): For scenarios involving parallel planes, the computation of the weighted midpoint involves fundamental mathematical operations, consequently resulting in a constant time complexity.
\item Defining a New Hyper-plane: The process of defining a new hyper-plane entails $M$ iterations over the corresponding norm vector, thereby giving rise to a time complexity of O(M), where $M$ is the number of dimensions.
\end{enumerate}

OLR-W\kern-0.25em A\kern-0.25em A integrates the aforementioned procedures without adaptive optimization mechanism comprising four core components: (1) drift detection, (2) drift magnitude quantification, (3) hyperparameter optimization, and (4) hyperparameter inference for adaptation. These components are executed sequentially for each incoming mini-batch and are succinctly outlined as follows:

\begin{enumerate}
    \item KPI Computation: The evaluation of performance metrics such as $R^2$ and MSE typically necessitates computing predictions of the form $\hat{y}_i = \mathbf{x}_i^\top \mathbf{w}, \text{ for } i = 1, \dots, K$, which involves a total computational complexity of $O(K M)$.

    \item KPI-Win Insertion: The computed KPIs are appended to the KPI-Win with a constant time complexity of $O(1)$.

    \item Statistical Measures: This step involves computing the mean $\mu$ and standard deviation $\sigma$ of the KPI-Win of length $|L|$, which requires $O(|L|)$ time. Additionally, the calculation of the threshold, as well as the lower and upper limits, involves standard operations with constant time complexity $O(1)$.

    \item Drift Magnitude Quantification: The deviation between the current KPI and the historical mean is calculated with a constant time complexity of $O(1)$.

    \item Scale Map: The interval between the KPI-Win mean and the lower or upper bound is partitioned into a fixed number of equal-width regions. Each region is then mapped to a corresponding subinterval within the full hyperparameter range, enabling scaled adaptation based on the degree of deviation. This procedure involves constant-time arithmetic and indexing, yielding a time complexity of $O(1)$.

    \item Hyperparameter Inference: Based on the position of the current KPI within the scale map, this step identifies and returns the corresponding hyperparameter value. The operation has a time complexity of $O(|L|)$.

    \item Model Adjustment via Inferred Hyperparameter: This step leverages all short-lived information from the current mini-batch, re-executes the EWMA update using the newly inferred hyperparameter and define the updated hyperplane, with an overall time complexity bounded by $O(M)$.
\end{enumerate}

In summary, the optimization mechanism in OLR-W\kern-0.25em A\kern-0.25em A has a time complexity bounded by $O(KM)$ and does not alter the overall complexity of OLR-W\kern-0.25em A, which remains dominated by the pseudo-inverse operation with a complexity of $O(NM^2)$.

\subsubsection*{Memory Complexity}

The memory footprint of OLR-W\kern-0.25em A\kern-0.25em A is primarily determined by the model parameters and the KPI-Win buffer used for performance monitoring and drift analysis. The model maintains a weight vector \(\mathbf{w} \in \mathbb{R}^M\), where \(M\) is the number of input features, requiring \(O(M)\) space. In addition, the KPI-Win buffer stores the most recent \(|L|\) performance entries, where each entry contains \(s\) scalar KPIs. Here, \(s\) denotes the number of evaluation metrics tracked per mini-batch, such as the mean squared error (MSE) and the coefficient of determination (\(R^2\)), which corresponds to \(s = 2\) in typical configurations. This results in a total memory complexity of \(O(M + |L| \cdot s)\), which remains minimal and scales well in streaming environments, making the approach 
suitable for deployment in high-throughput streaming environments.


\subsection{Evaluation Metric}

The coefficient of determination ($R^2$) is preferred in regression scenarios because it offers bounded and interpretable values, making it more informative than other common metrics such as MSE, MAE, RMSE, and MAPE~\cite{chicco2021coefficient}. Specifically, $R^2$ values range from negative infinity to 1, with values close to 1 indicating a strong predictive model. Unlike MSE or RMSE, which are unbounded and sensitive to the scale of the target variable, $R^2$ provides a normalized measure of fit that reflects the proportion of variance in the dependent variable explained by the independent variables. This scale-invariance allows for easier comparison of model performance across datasets and domains~\cite{chicco2021coefficient}. Moreover, $R^2$ generates high scores only when the majority of ground truth elements are predicted accurately, reflecting both the explained variance and the distribution of the data. As the most widely used measure of model fit in linear regression~\cite{montgomery2021introduction}, $R^2$ is often the default metric in statistical software and machine learning libraries. Its interpretability, statistical grounding~\cite{nagelkerke1991note}, and widespread adoption across scientific disciplines justify its recommendation as a standard evaluation criterion for regression models.

However, in one of our experiments, which was targeted to compute the average rank of models across multiple datasets in preparation for potential statistical significance testing, we selected the Mean Squared Error (MSE) as the evaluation metric. In addition to being the default cost function in regression \cite{hastie2009elements}, MSE was chosen due to its compatibility with parametric statistical tests that rely on assumptions of normally distributed errors, as commonly assumed in classical regression settings~\cite{montgomery2021introduction}. Unlike the coefficient of determination ($R^2$), which is bounded in $(-\infty, 1]$ and relative to the variance of the response variable, MSE serves as an absolute measure of predictive error. It consistently penalizes large deviations and preserves cardinal information about model performance~\cite{chicco2021coefficient}. These properties make MSE especially suitable for reliably ranking models across heterogeneous datasets, particularly when differences in performance are subtle or when dataset scales vary.

\section{Hyperparameters Sensitivity Analysis}

In this section we study the sensitivity of OLR-W\kern-0.25em A to its hyperparameter $(\alpha)$ across three distinct usage types of OLR-W\kern-0.25em A: Normal Online Linear Regression, Time-Based Adversarial Scenarios, and finally Confidence-Based Adversarial Scenarios.

\subsubsection*{Normal Linear Regression Scenarios}

Table \ref{tab:ablationstudy_normallinearregression} presents the performance results for each normal linear regression dataset, across diverse hyperparameter configurations. The findings indicate that OLR-W\kern-0.25em A demonstrates minimal sensitivity to the hyperparameter $(\alpha)$ within normal linear regression scenarios.

\begin{table*}[h] 
\captionsetup{justification=centering}
\caption{OLR-W\kern-0.25em A Sensitivity to Hyperparameters on Normal Regression\\
{\scriptsize Summary of Performance Measures using $R^2$ on the Last Iteration}}
\centering
\renewcommand{\theadalign}{cc} 
\renewcommand{\theadfont}{\small} 
\setlength{\tabcolsep}{4pt} 
\begin{adjustbox}{width=\textwidth} 
\begin{tabular}{|c|*{5}{c|}}
\hline
\multirow{2}{*}{} & \multicolumn{5}{c|}{\thead{Algorithms}} \\ \hline
{\thead{Datasets}}&
\thead{Equal Weights \\ $\alpha=0.5$} & 
\thead{Favor New Data\\$\alpha=.9$} & 
\thead{Radically Favor New data\\$\alpha=.995$} & 
\thead{Favor Existing Data \\$\alpha=.1$} & 
\thead{Radically Favor Existing Data\\ $\alpha=.005$} \\ \hline
\thead{DS1 \cite{OLR-WA_GIT}}&\makecell{0.97423}& \makecell{0.97154} & \makecell{0.97064} & \makecell{0.97612} & \makecell{0.97590} \\ \hline
\thead{DS2 \cite{OLR-WA_GIT}}& \makecell{0.98260} & \makecell{0.98027}& \makecell{0.97947 }& \makecell{0.98402}& \makecell{0.98389} \\ \hline
\thead{DS3 \cite{OLR-WA_GIT}}& \makecell{0.98204} & \makecell{0.98018}& \makecell{0.97952}& \makecell{0.98193}& \makecell{0.97792} \\ \hline
\thead{DS4 \cite{OLR-WA_GIT}}& \makecell{0.92482} & \makecell{0.91634}& \makecell{0.91326}& \makecell{0.92911}& \makecell{0.92144} \\ \hline
\thead{MCPD \cite{MCPD_DS}}& \makecell{0.73125} & \makecell{0.69246}& \makecell{0.67260}& \makecell{0.74188}& \makecell{0.72409}  \\ \hline
\thead{1KC\cite{1KC_DS}}& \makecell{0.90773} & \makecell{0.90545}& \makecell{0.90449}& \makecell{0.90646}& \makecell{0.90474} \\ \hline
\thead{KCHSD \cite{KCHS_DS}}& \makecell{0.54048}  &\makecell{0.45209} &\makecell{0.41781} & \makecell{0.57392}& \makecell{0.57453}  \\ \hline
\thead{CCPP \cite{CCPP_DS}}&  \makecell{0.92202}&\makecell{0.90869} & \makecell{0.90279}& \makecell{0.92785}& \makecell{0.92841} \\ \hline
\end{tabular}
\end{adjustbox}
\label{tab:ablationstudy_normallinearregression}
\end{table*}

\begin{table*}[h] 
\captionsetup{justification=centering}
\caption{OLR-W\kern-0.25em A Sensitivity to Hyperparameters on Adversarial Scenarios\\
{\scriptsize DS5, DS6 for Time-Based, and DS7, DS8 for Confidence-Based}\\
{\scriptsize Summary of Performance Measures using $R^2$ on the Last Iteration}}
\centering
\renewcommand{\theadalign}{cc} 
\renewcommand{\theadfont}{\small} 
\setlength{\tabcolsep}{4pt} 
\begin{adjustbox}{width=\textwidth} 
\begin{tabular}{|c|*{5}{c|}}
\hline
\multirow{2}{*}{} & \multicolumn{5}{c|}{\thead{Algorithms}} \\ \hline
{\thead{Datasets}}&
\thead{Equal Weights \\ $\alpha=0.5$} & 
 \thead{Favor New Data\\$\alpha=.9$} & 
 \thead{Radically Favor New data\\ $\alpha=.995$} & 
 \thead{Favor Existing Data \\$\alpha=.1$} & \thead{Radiclly Favor Existing Data\\$\alpha=.005$} \\ \hline
 \thead{DS5 \cite{OLR-WA_GIT}}&\makecell{0.98601}& \makecell{0.98513} & \makecell{0.98484} & \makecell{N/A} & \makecell{N/A} \\ \hline
\thead{DS6 \cite{OLR-WA_GIT}}& \makecell{0.60221} & \makecell{0.93739}& \makecell{0.93535}& \makecell{N/A}& \makecell{N/A} \\ \hline
\thead{DS7 \cite{OLR-WA_GIT}}& \makecell{N/A} & \makecell{N/A}& \makecell{N/A}& \makecell{0.07826}& \makecell{0.97815} \\ \hline
\thead{DS8 \cite{OLR-WA_GIT}}& \makecell{N/A} & \makecell{N/A}& \makecell{N/A}& \makecell{0.58033}& \makecell{0.93191} \\ \hline
\end{tabular}
\end{adjustbox}
\label{tab:ablationstudy_adversarial}
\end{table*}

\subsubsection*{Adversarial Scenarios}

Table \ref{tab:ablationstudy_adversarial} presents the performance results for DS5, and DS6 for Time based adversarial Scenarios, and DS7, and DS8 for Confidence-Based Adversarial Scenarios, cross diverse hyperparameter configurations. The findings indicate that OLR-W\kern-0.25em A demonstrates high sensitivity to the hyperparameter $(\alpha)$ within adversarial scenarios.

However, the conducted ablation study evaluates results on the test data after processing all data points, which is important but may not capture the model's sensitivity to internal fluctuations within each encountered mini-batch. To address this, the experimental section includes a dedicated part for concept drift experiments during data streams.
\section{Experiments}
This section presents experiments on synthetic and real public datasets, employing a rigorous methodology for robustness. To reduce randomness, results are averaged over 5 random seeds and evaluated using 5-fold cross-validation. Each reported \( R^2 \) value is based on 25 experiments with different data splits, ensuring stability and consistency.

In our experimentation, specific guidelines were diligently adhered to, guaranteeing an impartial evaluation of all models and preventing any form of bias.

\begin{enumerate}
    \item \small Initialization - During the initialization phase, all models' weights were set to an array of zeros, establishing a uniform and consistent starting point for each model.
    \item \small Feature Engineering - In the process of feature engineering, caution was exercised to limit any potential undue impact resulting from feature manipulation. Consequently, minimal feature engineering techniques, such as normalization and one-hot encoding, were employed to ensure the preservation of data integrity and maintain a balanced evaluation of the models.
    \item \small Hyperparameters Tuning - Hyperparameters were meticulously fine-tuned for each model and experiment, taking into account factors such as data set size and the number of features. The optimal outcomes for each model were carefully identified and subsequently reported.
    \item \small Mini-Batch Size (K) - For models utilizing mini-batches, such as MBGD and OLR-W\kern-0.25em A, the mini-batch size (K) was determined using the formula provided in Equation \ref{eq:olr_wa_batch_size}. The mini-batch size was set to either four times the number of dimensions in the dataset or a user-defined value, whichever was larger. In our experiments, the user-defined value was deliberately kept low to prioritize the formula-based size. This approach was adopted to strike a balance between efficient computation and model performance during our experimentation.
    \item \small Epochs Selection: 
    \begin{enumerate}
        \item \small For SGD, ORR, OLR online models, the appropriate hyperparameter was determined using the formula \(N \times \mathbb{Z}^{+}\), where $\mathbb{Z}^{+}$ is a Natural Number $>1$.
        \item \small For mini-batch models, the formula \(\frac{N}{K} \times \mathbb{Z}^{+}\) was utilized.
        Here, \(N\) refers to the number of training samples, \(M\) represents the data-set dimensions, and $\mathbb{Z}^{+}$ is a natural number. $\mathbb{Z}^{+}$ was systematically increased until no further improvement in the coefficient of determination ($R^2$) was observed.
    \end{enumerate}
\end{enumerate}

\begin{equation}
\label{eq:olr_wa_batch_size}
\text{K (Mini-Batch Size)} = \max\left(U, (M \times \mathbb{Z}^{+})\right)
\end{equation}


\begin{table*}[t]
\captionsetup{justification=centering}
 \caption{Summary of the Main Properties of the Datasets Considered.\\
        {\scriptsize Used for 1st Experiment: Performance Analysis on Normal Regression Scenarios}}
\centering
\renewcommand{\theadalign}{cc} 
\renewcommand{\theadfont}{\small} 
\setlength{\tabcolsep}{4pt} 
\begin{adjustbox}{width=\textwidth} 
\begin{tabular}{|c|*{7}{c|}}
\hline
\thead{Dataset} & \thead{Type} & \thead{Datapoints} & \thead{Dimensions} & \thead{Noise} & \thead{Training (5-Folds)} & \thead{Testing (5-Folds)} & \thead{Measurement\\Focus}\\
\hline 
DS1 \cite{OLR-WA_GIT} & Synthetic & 1,000 & 3 & 10 & 800 & 200 & Normal Regression \\
\hline
DS2 \cite{OLR-WA_GIT} & Synthetic & 10,000 & 20 & 20 & 8,000 & 2,000 &Normal Regression\\
\hline
DS3 \cite{OLR-WA_GIT} & Synthetic & 10,000 & 200 & 25 & 8,000 & 2,000 &Normal Regression\\
\hline
DS4 \cite{OLR-WA_GIT} & Synthetic & 50,000 & 500 & 50 & 40,000 & 10,000 &Normal Regression\\
\hline
MCPD \cite{MCPD_DS} & Real & 1,338 & 7 & \makecell{N/A} & 1,070 & 267 &Normal Regression\\
\hline
1KC \cite{1KC_DS} & Real & 1,000 & 5 & \makecell{N/A} & 800 & 200 &Normal Regression\\
\hline
KCHSD \cite{KCHS_DS} & Real & 21,613 & 21 & \makecell{N/A} & 17,290 & 4,322 &Normal Regression\\
\hline
CCPP \cite{CCPP_DS} & Real & 9,568 & 5 & \makecell{N/A} & 7,654 & 1,914&Normal Regression \\
\hline
DS5 \cite{OLR-WA_GIT} & Synthetic & 5,000 & 20 & 20 & 4,500 & 500 & Time-Based \\
\hline
DS6 \cite{OLR-WA_GIT} & Synthetic & 10,000 & 200 & 40 & 9,000 & 1,000 & Time-Based \\
\hline
DS7 \cite{OLR-WA_GIT} & Synthetic & 5,000 & 20 & 20 & 4,500 & 500 & Confidence-Based \\
\hline
DS8 \cite{OLR-WA_GIT} & Synthetic & 10,000 & 200 & 40 & 9,000 & 1,000 & Confidence-Based \\
\hline
DS9 \cite{OLR-WA_GIT} & Synthetic & 1,000 & 2 & 20 & 800 & 200 & Convergence \\
\hline
DS10 \cite{OLR-WA_GIT} & Synthetic & 1,000 & 2 & 40 & 800 & 200 & Convergence \\
\hline
\end{tabular}
\end{adjustbox}
\label{tab:datasets-properties}
\end{table*}

\subsection{Performance Analysis on Stationary Linear Regression Scenarios}
A comparative analysis was conducted to evaluate the performance of different online regression models, employing synthetic and real-world publicly available data-sets. This experiment, utilizing OLR-W\kern-0.25em A, was conducted under stationary regression settings where no concept drift is present. The evaluation utilized 5-fold cross-validation and seed averaging, with the $R^2$ metric measured at the end of the last iteration. To provide a concise overview of our experimental setup, Table \ref{tab:datasets-properties} presents key properties of the specific data-sets employed in this study.

\begin{table*}[!ht]
\captionsetup{justification=centering}
\caption{OLR-W\kern-0.25em A Performance Analysis on Normal Regression Scenarios \\
{\scriptsize Summary of Performance Measures using $R^2$ on the Last Iteration \\ 
\scriptsize $t_s$: time in seconds, E: epochs, $\mathbb{Z}^{+}$: Natural Number $\geq$ 1}}
\centering
\renewcommand{\theadalign}{cc} 
\renewcommand{\theadfont}{\small} 
\setlength{\tabcolsep}{4pt} 
\begin{adjustbox}{width=\textwidth} 
\begin{tabular}{|c|*{9}{c|}}
\hline
\multirow{2}{*}{} & \multicolumn{9}{c|}{\thead{Algorithms}} \\ \hline
{\thead{Datasets}}&
\rotatebox[origin=c]{90}{\thead{\textbf{Batch Regression} \\(Pseudo-Inverse)}} & 
\rotatebox[origin=c]{90}{\thead{\textbf{SGD}\\ {\scriptsize $\eta=0.01$ [DS3, DS3 =0.01},\\ \scriptsize{ KCHSD 0.0001]}, \\ {\scriptsize E=N $\times$ $(\mathbb{Z}^{+}=2)$ [1KC $\mathbb{Z}^{+}=3$,}\\{\scriptsize CCPP $\mathbb{Z}^{+}=5$]}}} & 
\rotatebox[origin=c]{90}{\thead{\textbf{MBGD} \\ {\scriptsize $\eta=0.01$ [DS4= 0.001],}\\{\scriptsize K=M $\times (\mathbb{Z}^{+} = 5)$,}\\{\scriptsize $E=\frac{N}{K} \times (\mathbb{Z}^{+} =10)$ [DS1 $\mathbb{Z}^{+}$=5,}\\{\scriptsize DS3 $\mathbb{Z}^{+}$=100, KCHSD $\mathbb{Z}^{+}$=20,}\\{\scriptsize 1KC, CCPP $\mathbb{Z}^{+}$=40]}}} & 
\rotatebox[origin=c]{90}{\thead{\textbf{LMS}\\ {\scriptsize $\eta=0.01$[1KC=0.001},\\ \scriptsize{KCHSD=0.0001]}}} & 
\rotatebox[origin=c]{90}{\thead{\textbf{ORR}\\ {\scriptsize $\eta=0.01$ [DS3, DS4, KCHSD,}\\ \scriptsize{MCPD=0.001], E=N $\times (\mathbb{Z}^{+}=2)$}\\{\scriptsize [1KC $\mathbb{Z}^{+}$=3, KCHSD, CCPP}\\ \scriptsize{ $\mathbb{Z}^{+}=5$], $\lambda=0.1$ [1KC,}\\{\scriptsize [MCPD,KCHSD,CCPP=0.001]}}} & 
\rotatebox[origin=c]{90}{\thead{\textbf{OLR}\\ {\scriptsize $\eta=0.01$ [DS3, DS4, MCPD, KCHSD}\\{\scriptsize = 0.001], E=N $\times (\mathbb{Z}^{+}=2)$}\\{\scriptsize [1KC $\mathbb{Z}^{+}$=3, KCHSD, CCPP $\mathbb{Z}^{+}=5$], }\\{\scriptsize $\lambda=0.1$ [MCPD, CCPP=0.01],}\\ \scriptsize{[1KC, KCHSD=$0.001$],[KCHSD=$0.0001$]}}} & 
\rotatebox[origin=c]{90}{\thead{\textbf{RLS} \\ {\scriptsize $\lambda=.99$, $\delta=0.01$}}} &                        
\rotatebox[origin=c]{90}{\thead{\textbf{PA}\\ {\scriptsize $C=0.01$, $\epsilon=0.01$},\\ \scriptsize{ [DS1, DS2, DS3}\\ \scriptsize{$C=0.1$, $\epsilon=0.1$]}}} & 
\rotatebox[origin=c]{90}{\thead{\textbf{OLR-W\kern-0.25em A}\\ {\scriptsize  $\alpha=.5$}\\{\scriptsize [KCHSD  $\alpha=.1$],}\\{\scriptsize $BK=N \times (\mathbb{Z}^{+} = 0.1)$,}\\{\scriptsize K=M $\times$ $(\mathbb{Z}^{+} = 5)$}}} \\ \hline
\thead{DS1 \cite{OLR-WA_GIT}}&\makecell{$r^{\lower0.01ex\hbox{\raisebox{-1.5pt}{\tiny$2$}}}
$: 0.97637 \\ $t_s$: 0.00121s }& \makecell{$r^{\lower0.01ex\hbox{\raisebox{-1.5pt}{\tiny$2$}}}
$: 0.97561 \\ $t_s$: 0.01391s} & \makecell{$r^{\lower0.01ex\hbox{\raisebox{-1.5pt}{\tiny$2$}}} $: 0.97580  \\ $t_s$: 0.00434s } & \makecell{$r^{\lower0.01ex\hbox{\raisebox{-1.5pt}{\tiny$2$}}} $: 0.97616 \\ $t_s$: 0.00669s} & \makecell{$r^{\lower0.01ex\hbox{\raisebox{-1.5pt}{\tiny$2$}}} $: 0.96760 \\ $t_s$: 0.01958s }& \makecell{$r^{\lower0.01ex\hbox{\raisebox{-1.5pt}{\tiny$2$}}} $: 0.97565 \\ $t_s$: 0.02026s}& \makecell{$r^{\lower0.01ex\hbox{\raisebox{-1.5pt}{\tiny$2$}}} $: 0.97624 \\ $t_s$: 0.01073s} & \makecell{$r^{\lower0.01ex\hbox{\raisebox{-1.5pt}{\tiny$2$}}} $: 0.97378 \\ $t_s$: 0.00866s} & \makecell{$r^{\lower0.01ex\hbox{\raisebox{-1.5pt}{\tiny$2$}}} $: 0.97423 \\ $t_s$: 0.01185s}   \\ \hline
\thead{DS2 \cite{OLR-WA_GIT}}& \makecell{$r^{\lower0.01ex\hbox{\raisebox{-1.5pt}{\tiny$2$}}} $: 0.98418 \\ $t_s$: 0.01431s  } & \makecell{$r^{\lower0.01ex\hbox{\raisebox{-1.5pt}{\tiny$2$}}} $: 0.98231 \\ $t_s$: 0.14151s }& \makecell{$r^{\lower0.01ex\hbox{\raisebox{-1.5pt}{\tiny$2$}}} $: 0.98384 \\ $t_s$: 0.01453s}& \makecell{$r^{\lower0.01ex\hbox{\raisebox{-1.5pt}{\tiny$2$}}} $: 0.98243 \\ $t_s$: 0.06802s}& \makecell{$r^{\lower0.01ex\hbox{\raisebox{-1.5pt}{\tiny$2$}}} $: 0.96987 \\ $t_s$: 0.19516s}& \makecell{$r^{\lower0.01ex\hbox{\raisebox{-1.5pt}{\tiny$2$}}} $: 0.97959 \\ $t_s$: 0.20398s}& \makecell{$r^{\lower0.01ex\hbox{\raisebox{-1.5pt}{\tiny$2$}}} $: 0.98268 \\ $t_s$: 0.12351s}&  \makecell{$r^{\lower0.01ex\hbox{\raisebox{-1.5pt}{\tiny$2$}}} $: 0.97642 \\ $t_s$: 0.05957s}& \makecell{$r^{\lower0.01ex\hbox{\raisebox{-1.5pt}{\tiny$2$}}} $: 0.98260 \\ $t_s$: 0.03571s}  \\ \hline
\thead{DS3 \cite{OLR-WA_GIT}}&  \makecell{$r^{\lower0.01ex\hbox{\raisebox{-1.5pt}{\tiny$2$}}} $: 0.98299 \\ $t_s$: 0.08145s} & \makecell{$r^{\lower0.01ex\hbox{\raisebox{-1.5pt}{\tiny$2$}}} $: 0.98116 \\ $t_s$: 0.15012s }& \makecell{$r^{\lower0.01ex\hbox{\raisebox{-1.5pt}{\tiny$2$}}} $: 0.98290 \\ $t_s$: 0.24452s}& \makecell{$r^{\lower0.01ex\hbox{\raisebox{-1.5pt}{\tiny$2$}}} $: 0.98161 \\ $t_s$: 0.07244s }& \makecell{$r^{\lower0.01ex\hbox{\raisebox{-1.5pt}{\tiny$2$}}} $: 0.96901 \\ $t_s$: 0.18893s }& \makecell{$r^{\lower0.01ex\hbox{\raisebox{-1.5pt}{\tiny$2$}}} $: 0.97933 \\ $t_s$: 0.20211s}& \makecell{$r^{\lower0.01ex\hbox{\raisebox{-1.5pt}{\tiny$2$}}} $: 0.96001 \\ $t_s$: 5.14062s }& \makecell{$r^{\lower0.01ex\hbox{\raisebox{-1.5pt}{\tiny$2$}}} $: 0.96705 \\ $t_s$: 0.09319s }& \makecell{$r^{\lower0.01ex\hbox{\raisebox{-1.5pt}{\tiny$2$}}} $: 0.98204 \\ $t_s$: 0.39293s}  \\ \hline
\thead{DS4 \cite{OLR-WA_GIT}}& \makecell{$r^{\lower0.01ex\hbox{\raisebox{-1.5pt}{\tiny$2$}}} $: 0.92973 \\ $t_s$: 49.89187s} & \makecell{$r^{\lower0.01ex\hbox{\raisebox{-1.5pt}{\tiny$2$}}} $:0.90663   \\ $t_s$: 5.95597s }& \makecell{$r^{\lower0.01ex\hbox{\raisebox{-1.5pt}{\tiny$2$}}} $: 0.92964 \\ $t_s$: 156.83463s}& \makecell{$r^{\lower0.01ex\hbox{\raisebox{-1.5pt}{\tiny$2$}}} $: 0.90715 \\ $t_s$: 5.02476s}& \makecell{$r^{\lower0.01ex\hbox{\raisebox{-1.5pt}{\tiny$2$}}} $: 0.85535\\ $t_s$: 1.92488s }& \makecell{$r^{\lower0.01ex\hbox{\raisebox{-1.5pt}{\tiny$2$}}} $: 0.86412 \\ $t_s$: 3.93231s}& \makecell{$r^{\lower0.01ex\hbox{\raisebox{-1.5pt}{\tiny$2$}}} $: 0.62667 \\ $t_s$: 1907.88762s}& \makecell{$r^{\lower0.01ex\hbox{\raisebox{-1.5pt}{\tiny$2$}}} $: 0.87255 \\ $t_s$: 3.80154s }& \makecell{$r^{\lower0.01ex\hbox{\raisebox{-1.5pt}{\tiny$2$}}} $: 0.92464 \\ $t_s$: 8.08490s}  \\ \hline
\thead{MCPD \cite{MCPD_DS}}& \makecell{$r^{\lower0.01ex\hbox{\raisebox{-1.5pt}{\tiny$2$}}} $: 0.74321 \\ $t_s$: 0.00332s} & \makecell{$r^{\lower0.01ex\hbox{\raisebox{-1.5pt}{\tiny$2$}}} $: 0.73092 \\ $t_s$: 0.02735s}& \makecell{$r^{\lower0.01ex\hbox{\raisebox{-1.5pt}{\tiny$2$}}} $: 0.73007 \\ $t_s$: 0.00905s}& \makecell{$r^{\lower0.01ex\hbox{\raisebox{-1.5pt}{\tiny$2$}}} $: 0.73637 \\ $t_s$: 0.01364s }& \makecell{$r^{\lower0.01ex\hbox{\raisebox{-1.5pt}{\tiny$2$}}} $: 0.74070 \\ $t_s$: 0.02973s}& \makecell{$r^{\lower0.01ex\hbox{\raisebox{-1.5pt}{\tiny$2$}}} $: 0.74217 \\ $t_s$: 0.02825s}& \makecell{$r^{\lower0.01ex\hbox{\raisebox{-1.5pt}{\tiny$2$}}} $: 0.73830 \\ $t_s$: 0.02185s}&  \makecell{$r^{\lower0.01ex\hbox{\raisebox{-1.5pt}{\tiny$2$}}} $: 0.72893 \\ $t_s$: 0.00650s} & \makecell{$r^{\lower0.01ex\hbox{\raisebox{-1.5pt}{\tiny$2$}}} $: 0.74080 \\ $t_s$: 0.00358s}  \\ \hline
\thead{1KC\cite{1KC_DS}}& \makecell{$r^{\lower0.01ex\hbox{\raisebox{-1.5pt}{\tiny$2$}}} $: 0.93615 \\ $t_s$: 0.00111s} & \makecell{$r^{\lower0.01ex\hbox{\raisebox{-1.5pt}{\tiny$2$}}} $: 0.92930 \\ $t_s$: 0.01471s}& \makecell{$r^{\lower0.01ex\hbox{\raisebox{-1.5pt}{\tiny$2$}}} $: 0.91195 \\ $t_s$: 0.01702s }& \makecell{$r^{\lower0.01ex\hbox{\raisebox{-1.5pt}{\tiny$2$}}} $: 0.60772 \\ $t_s$: 0.00470s }& \makecell{$r^{\lower0.01ex\hbox{\raisebox{-1.5pt}{\tiny$2$}}} $: 0.90294 \\ $t_s$: 0.02493s}& \makecell{$r^{\lower0.01ex\hbox{\raisebox{-1.5pt}{\tiny$2$}}} $: 0.90322 \\ $t_s$: 0.02582s}& \makecell{$r^{\lower0.01ex\hbox{\raisebox{-1.5pt}{\tiny$2$}}} $: 0.85503  \\ $t_s$: 0.00807s}& \makecell{$r^{\lower0.01ex\hbox{\raisebox{-1.5pt}{\tiny$2$}}} $: 0.78565 \\ $t_s$: 0.00479s }& \makecell{$r^{\lower0.01ex\hbox{\raisebox{-1.5pt}{\tiny$2$}}} $: 0.90773 \\ $t_s$: 0.00606s}  \\ \hline
\thead{KCHSD \cite{KCHS_DS}}& \makecell{$r^{\lower0.01ex\hbox{\raisebox{-1.5pt}{\tiny$2$}}} $: 0.57859\\ $t_s$: 0.02213s}  &\makecell{$r^{\lower0.01ex\hbox{\raisebox{-1.5pt}{\tiny$2$}}} $: 0.57277 \\ $t_s$: 0.31353s} &\makecell{$r^{\lower0.01ex\hbox{\raisebox{-1.5pt}{\tiny$2$}}} $: 0.57431 \\ $t_s$: 0.17558s} & \makecell{$r^{\lower0.01ex\hbox{\raisebox{-1.5pt}{\tiny$2$}}} $: 0.56344 \\ $t_s$: 0.11620s}& \makecell{$r^{\lower0.01ex\hbox{\raisebox{-1.5pt}{\tiny$2$}}} $: 0.56432 \\ $t_s$: 0.88060s }& \makecell{$r^{\lower0.01ex\hbox{\raisebox{-1.5pt}{\tiny$2$}}} $: 0.57040\\ $t_s$: 0.95015s}& \makecell{$r^{\lower0.01ex\hbox{\raisebox{-1.5pt}{\tiny$2$}}} $: 0.48088 \\ $t_s$: 0.18357s }&  \makecell{$r^{\lower0.01ex\hbox{\raisebox{-1.5pt}{\tiny$2$}}} $: 0.53533 \\ $t_s$: 0.12020s }& \makecell{$r^{\lower0.01ex\hbox{\raisebox{-1.5pt}{\tiny$2$}}} $: 0.57395 \\ $t_s$: 0.08367s}  \\ \hline
\thead{CCPP \cite{CCPP_DS}}&  \makecell{$r^{\lower0.01ex\hbox{\raisebox{-1.5pt}{\tiny$2$}}} $: 0.92855\\ $t_s$: 0.00807s}&\makecell{$r^{\lower0.01ex\hbox{\raisebox{-1.5pt}{\tiny$2$}}} $: 0.91838\\ $t_s$: 0.22912s} & \makecell{$r^{\lower0.01ex\hbox{\raisebox{-1.5pt}{\tiny$2$}}} $: 0.91989\\$t_s$: 0.29177s}& \makecell{$r^{\lower0.01ex\hbox{\raisebox{-1.5pt}{\tiny$2$}}} $: 0.89785\\ $t_s$: 0.04680s}& \makecell{$r^{\lower0.01ex\hbox{\raisebox{-1.5pt}{\tiny$2$}}} $: 0.92550\\ $t_s$: 0.39269s}& \makecell{$r^{\lower0.01ex\hbox{\raisebox{-1.5pt}{\tiny$2$}}} $: 0.92611\\ $t_s$: 0.58023s}&\makecell{$r^{\lower0.01ex\hbox{\raisebox{-1.5pt}{\tiny$2$}}} $: 0.66197\\ $t_s$: 0.07742s}& \makecell{$r^{\lower0.01ex\hbox{\raisebox{-1.5pt}{\tiny$2$}}} $: 0.66327\\ $t_s$: 0.04727s}& \makecell{$r^{\lower0.01ex\hbox{\raisebox{-1.5pt}{\tiny$2$}}} $: 0.92202\\ $t_s$: 0.05153s}  \\ \hline
\end{tabular}
\end{adjustbox}
\label{tab:algorithm-performance}
\end{table*}

\textit{Performance Evaluation }
Table \ref{tab:algorithm-performance} presents the performance results for each data-set, along with the hyperparameters utilized for each model. The inclusion of these hyperparameters enhances transparency and facilitates a comprehensive analysis of the models' performance. Below are some observed insights about the performance evaluation for this experiment:
\begin{enumerate}
    \item \small The batch model, depicted in the first column, serves as a benchmark for evaluating performance and sets the standard for all online models.
    \item \small OLR-W\kern-0.25em A performance is remarkably significant and very close to the batch model, the difference is very slight and almost in the 3'd digit after the decimal point across all data-sets, except the 1KC data-set, where the batch model performance is 0.93615 while OLR-W\kern-0.25em A is 0.90773.
    \item \small RLS performance significantly deteriorates on the more challenging DS4 data-set, where it achieves an $R^2$ value of approximately 0.62667. Additionally, it demonstrates a low $R^2$ score of 0.66197 on the CCPP data-set.
    \item \small PA has demonstrated its proficiency in capturing underlying patterns and relationships effectively, as evidenced by competitive $R^2$ values across various datasets. However, its performance on the 1KC dataset appears to be somewhat limited, with an $R^2$ value of approximately 0.78565, in contrast to the batch model's 0.93615. Similarly, on the CCPP dataset, PA achieved an $R^2$ value of 0.66327, notably lower than the batch model's 0.92855. 

    \item \small The Widrow-Hoff algorithm exhibits good performance in most of the data-sets, although it experienced some degradation with an $R^2$ value of 0.60772 in the 1KC data-set, whereas the batch model $R^2$ value is 0.93615.
\end{enumerate}

In conclusion, the application of OLR-W\kern-0.25em A with its default hyperparameter setting of $\alpha = 0.5$ consistently delivered top-tier performance across all evaluated data-sets.

\subsection{Performance Analysis on Adversarial Scenarios}

This experiment aims to understand the model's behavior under varying settings of the hyperparameter $\alpha$ in adversarial data environments. A high $\alpha$ value is used in Time-Based scenarios to enable rapid adaptation to recent patterns, while a low $\alpha$ is used in Confidence-Based scenarios to emphasize stability and trust in the existing model. This setup employs OLR-W\kern-0.25em A, which involves manual tuning of $\alpha$, and highlights the need for dynamic adjustment, a feature introduced in OLR-W\kern-0.25em A\kern-0.25em A.

As summarized in Table \ref{tab:datasets-properties}, datasets DS5 through DS8 were used for the experiment. DS5 and DS6 correspond to time-based scenarios, where the model is expected to prioritize recent observations following a distributional shift. DS7 and DS8 represent confidence-based scenarios, evaluating the model's ability to maintain performance by relying on historically consistent patterns.

\begin{table*}[!ht]
\captionsetup{justification=centering}
\caption{OLR-W\kern-0.25em A Performance Analysis on Adversarial Scenarios\\
{\scriptsize  Summary of Performance Measures using $R^2$ on the Last Iteration\\ 
\scriptsize N/A: Minus $R^2$, $E$=Epochs, $\mathbb{Z}^{+}$: Natural Number $\geq$ 1}}
\centering
\renewcommand{\theadalign}{cc} 
\renewcommand{\theadfont}{\small} 
\setlength{\tabcolsep}{4pt} 
\begin{adjustbox}{width=\textwidth} 
\begin{tabular}{|c|*{9}{c|}}
\hline
\multirow{2}{*}{} & \multicolumn{9}{c|}{\thead{Algorithms}} \\ \hline
{\thead{Datasets}}
& \rotatebox[origin=c]{90}{\thead{\textbf{Batch Regression} \\(Pseudo-Inverse)}} & 
\rotatebox[origin=c]{90}{\thead{\textbf{SGD} \\ {\scriptsize $\eta=$[DS5, DS7 = 0.01,}\\{\scriptsize DS6, DS8=0.001],}\\{\scriptsize E=N $\times (\mathbb{Z}^{+} = 2)$}}} & 
\rotatebox[origin=c]{90}{\thead{\textbf{MBGD} \\ {\scriptsize $\eta=0.01$, K=M $\times (\mathbb{Z}^{+} = 5)$}\\{\scriptsize E=$\frac{N}{K} \times (\mathbb{Z}^{+} = 2)$}}} & 
\rotatebox[origin=c]{90}{\thead{\textbf{LMS}\\ {\scriptsize $\eta=$ [DS5, DS7 = 0.01,}\\{\scriptsize DS6, DS8 = 0.001]}}} & 
\rotatebox[origin=c]{90}{\thead{\textbf{ORR}\\ {\scriptsize $\eta=$ [DS5, DS7 = 0.01,}\\{\scriptsize DS6, DS8 = 0.001]} \\{\scriptsize E = N $\times (\mathbb{Z}^{+} = 2)$,}\\{\scriptsize $\lambda = $[DS5, DS7 = 0.1,}\\{\scriptsize DS6, DS8 = 0.01]}}}
& \rotatebox[origin=c]{90}{\thead{\textbf{OLR}\\ {\scriptsize $\eta=$ [DS5, DS7 = 0.01,}\\{\scriptsize DS6, DS8 = 0.001]} \\{\scriptsize E = N $\times (\mathbb{Z}^{+} = 2)$,}\\{\scriptsize $\lambda = $[DS5, DS7 = 0.1,}\\{\scriptsize DS6, DS8 = 0.01]}}}
& \rotatebox[origin=c]{90}{\thead{\textbf{RLS} \\ {\scriptsize $\lambda$= [DS5, DS6 = .99,}\\{\scriptsize DS7, DS8 = .18], $\delta=0.01$}}}
& \rotatebox[origin=c]{90}{\thead{\textbf{PA}\\ {\scriptsize $C=0.1$, $\epsilon=0.1$}\\ \scriptsize{[DS6 $C=.01$, $\epsilon=.01$]}}}
& \rotatebox[origin=c]{90}{\thead{\textbf{OLR-W\kern-0.25em A}\\ {\scriptsize $\alpha=$ [DS5, DS6 = .95,}\\{\scriptsize DS7, DS8 = 0.005]}\\{\scriptsize $BK=N \times 0.1$,}\\{\scriptsize K=M $\times (\mathbb{Z}^{+} = 5) $ }}} \\ \hline
DS5 \cite{OLR-WA_GIT}& N/A & N/A & N/A & 0.98528 & N/A & N/A & 0.98546 & 0.97812 & 0.98498   \\ \hline
DS6 \cite{OLR-WA_GIT}& N/A & N/A & N/A & 0.93810 & N/A & N/A & 0.87134 & 0.91145 & 0.93634  \\ \hline
DS7 \cite{OLR-WA_GIT}&  N/A & N/A & N/A & N/A & N/A & N/A & N/A & N/A & 0.97815  \\ \hline
DS8 \cite{OLR-WA_GIT}& N/A & N/A & N/A & N/A & N/A & N/A & N/A & N/A & 0.93191  \\ \hline
\end{tabular}
\end{adjustbox}
\label{tab:algorithm-adv-performance}
\end{table*}

\textit{Performance Evaluation }
Table \ref{tab:algorithm-adv-performance} presents the performance results for each data-set, along with the hyperparameters utilized for each model.
Based on the experimental findings of DS5 and DS6, specifically designed for Time-Based scenarios, it becomes evident that the model's performance improves when it can effectively accommodate new data. Notably, OLR-W\kern-0.25em A, Widrow-Hoff, PA, and RLS algorithms have demonstrated remarkable outcomes in terms of their ability to adapt to dynamic changes in the data. In contrast, the performance results of other online models, namely SGD, MBGD, Online Ridge, Online Lasso, and even the standard batch model, were notably unsatisfactory within this specific scenario. 

The experimental results obtained from DS7 and DS8, which are tailored for Confidence-Based scenarios, highlight the significance of the model's capability to effectively incorporate previous data for improved performance. Particularly, the OLR-W\kern-0.25em A algorithm has exhibited impressive outcomes by consistently adapting to changes in data patterns while maintaining a conservative approach. Conversely, the performance of all other online models proved to be notably inadequate within this particular scenario.

The accomplishments of the Confidence-Based OLR-W\kern-0.25em A algorithm hold significant relevance across various scenarios, such as sentiment analysis tasks, where certain labeled data points have undergone verification by experts or trusted sources. In such cases, assigning higher weights to these points based on their elevated confidence levels can prove advantageous. A real case study of the applicability of linear regression on sentiment analysis is conducted by the authors in \cite{cakra2015stock}.
Another example, similarly consider the example of an Amazon store with an extensive product inventory and a daily influx of hundreds of new products. In this context, employing a Confidence-Based approach, which inherently favors the larger existing product pool, or products that have been selected for trustworthiness, can potentially enhance the model's prediction performance.

\begin{figure}[H]
    \centering
    \begin{subfigure}[t]{0.45\textwidth}
        \includegraphics[width=\linewidth]{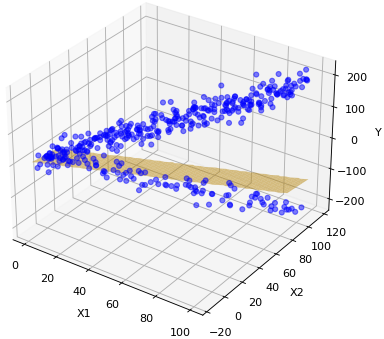}
        \caption{Time-Based Scenario}
        \label{fig:TimeBased}
    \end{subfigure}
    \hfill
    \begin{subfigure}[t]{0.45\textwidth}
        \includegraphics[width=\linewidth]{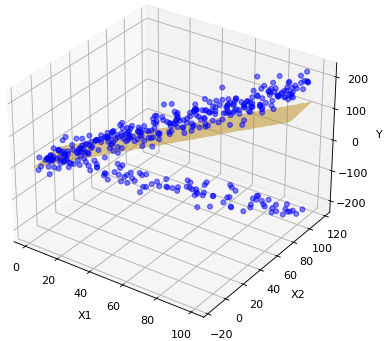}
        \caption{Confidence-Based Scenario}
        \label{fig:ConfidenceBased}
    \end{subfigure}
    \caption{Comparison of two adaptation strategies.}
    \label{fig:comparisonStrategies}
\end{figure}

As a key highlight of our visual experiment, Figures \ref{fig:TimeBased} and \ref{fig:ConfidenceBased} illustrate the behavior of OLR-W\kern-0.25em A in Time-Based and Confidence-Based scenarios. Figure \ref{fig:TimeBased} depicts a Time-Based adversarial setting, where OLR-W\kern-0.25em A, with its hyperparameter set to \(\alpha = 0.95\), successfully adapted to a newly introduced negatively correlated adversarial model. In contrast, Figure \ref{fig:ConfidenceBased} presents a confidence-based scenario, in which \(\alpha = 0.05\), fine-tuning enabled OLR-W\kern-0.25em A to preserve confidence in existing data, minimizing the impact of new data on its performance. These findings are significant because they provide valuable insight into automatic weight selection and adaptive hyperparameter tuning, particularly in the presence of concept drift or adversarial conditions.

\subsection{Convergence Analysis}

This experiment provides a comprehensive comparative analysis of the convergence behavior of online regression models, with a particular focus on OLR-W\kern-0.25em A. The evaluation examines OLR-W\kern-0.25em A's stability and efficiency relative to other models, employing the \( R^2 \) metric to assess predictive performance and Mean Squared Error (MSE) to analyze convergence trends, both monitored iteratively. To ensure robustness and generalizability, the analysis leverages 10 datasets with diverse characteristics, capturing a wide range of data distributions and complexities to thoroughly evaluate convergence dynamics.

Initially, reference datasets DS9 and DS10 \cite{OLR-WA_GIT}, characterized by reduced dimensionality, are employed to constrain information, thereby challenging model performance while enabling detailed convergence monitoring with smaller data batches. Subsequently, the models were assessed on a broader set of datasets with varying dimensionality and complexity to evaluate their generalization capabilities and convergence behavior under diverse data conditions.

\begin{figure}[H]
\centering
\includegraphics[width=\textwidth, height=0.34\textheight, keepaspectratio]{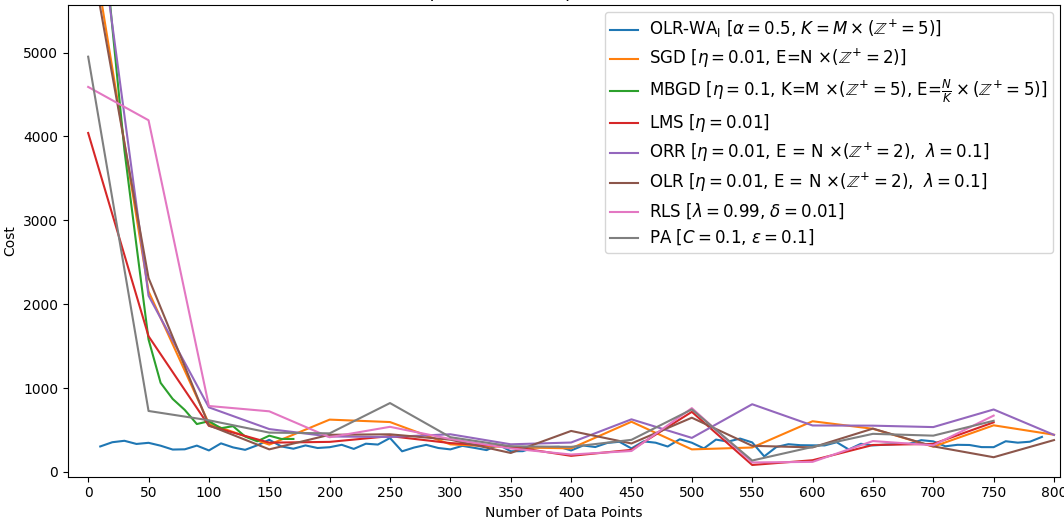}
\captionsetup{justification=centering}
\caption{Convergence Analysis\\
{\scriptsize Models Cost by Number of Data Points Considered on DS10 \\
for Multiple Seeds, 5-Fold Cross Validation}}
\label{fig:models_cost_convergence}
\end{figure}

Figure~\ref{fig:models_cost_convergence} illustrates the convergence behavior of various algorithms based on MSE using DS10~\cite{OLR-WA_GIT}. Most algorithms reach optimal performance around epoch 150. However, OLR-W\kern-0.25em A exhibits a distinct convergence pattern, starting with a significantly lower cost and continuously sustaining this diminished cost throughout. This early and consistent cost advantage establishes one of our key contributions in this work, showing that OLR-W\kern-0.25em A not only adapts quickly but also maintains superior performance over time. As shown in Table~\ref{tab:model-performance}, OLR-W\kern-0.25em A achieved the highest \( R^2 \) score of 0.86265 within the first 10 data points and sustained high \( R^2 \) values from the 90th data point onward. PA also demonstrated rapid convergence, stabilizing at a high \( R^2 \) by the 70th data point, while the majority of models converged between the 120th and 150th data points.

\begin{table*}[!ht]
\captionsetup{justification=centering}
\caption{Convergence Analysis \\
\scriptsize Models Performance by Number of Data Points Considered on DS9 \\for Multiple Seeds, 5 Folds Cross Validation\\
\scriptsize Hyperparameters are the same as Fig. \ref{fig:models_cost_convergence} }
\centering
\renewcommand{\theadalign}{cc} 
\renewcommand{\theadfont}{\small} 
\setlength{\tabcolsep}{4pt} 
\begin{adjustbox}{width=\textwidth} 
\begin{tabular}{|c|*{16}{c|}}
\hline
\multirow{2}{*}{} & \multicolumn{16}{c|}{\thead{Number of Training Data Points}} \\ \hline
{\thead{Model}}&
\textbf{10} & \textbf{20} & \textbf{30} & \textbf{40} & \textbf{50} & \textbf{60} & \textbf{70} & \textbf{80} & \textbf{90} & \textbf{100} & \textbf{110} & \textbf{120} & \textbf{130} & \textbf{140} & \textbf{150} & \textbf{Final}\\
\hline
\textbf{OLR-W\kern-0.25em A}         &0.86265&0.87026&0.87771&0.88329&0.88861&0.89288&0.89591&0.89875&0.90098&0.90161&0.90433&0.90615&0.90810&0.90986&0.91084&0.91293\\
\hline
\textbf{SGD}    &0.32402&0.50374&0.62607&0.72745&0.78288&0.82651&0.85521&0.87468&0.88738&0.89584&0.90132&0.90371&0.90599&0.90806&0.90799&0.91233\\
\hline
\textbf{MBGD}
&0.32226&0.51114&0.64383&0.74555&0.79964&0.83936&0.85881&0.87445&0.88487&0.89264&0.89909&0.90194&0.90535&0.90592&0.90829&0.91177\\
\hline
\textbf{LMS} &0.25216&0.48899&0.66102&0.72896&0.78882&0.82736&0.84889&0.86801&0.88491&0.89585&0.90026&0.90347&0.90625&0.90919&0.91053&0.91348\\
\hline
\textbf{ORR}   &0.28438&0.44982&0.59813&0.69093&0.74606&0.78599&0.81481&0.84090&0.85622&0.86723&0.88113&0.88760&0.89168&0.89530&0.89552&0.90296\\
\hline
\textbf{OLR}   &0.27546&0.51746&0.64119&0.72773&0.77723&0.81353&0.84392&0.86510&0.88134&0.89082&0.89761&0.90173&0.90476&0.90693&0.90769&0.91299\\
\hline
\textbf{RLS}     &0.13441&0.30085&0.43636&0.49871&0.57590&0.63265&0.67494&0.71707&0.75557&0.78608&0.80571&0.82775&0.83981&0.85533&0.87163&0.91513\\
\hline
\textbf{PA}            &0.76154&0.88878&0.89449&0.90396&0.86813&0.89807&0.90726&0.91309&0.90740&0.90570&0.90363&0.90456&0.90786&0.90941&0.90577&0.90626\\
\hline
\end{tabular}
\end{adjustbox}
\label{tab:model-performance}
\end{table*}


Finally, another thorough convergence experiment was conducted. Table \ref{table:models_performance_first_10_minibatches} presents the performance of the models by evaluating the MSE over the initial ten mini-batches. We utilized the same number of mini-batches ($K$) at each iteration ($t$) to ensure consistency and comparability across different models. The experiment evaluates the average rank of each model's performance across various datasets, assigning a rank of 1 to the top-performing model. Among the models assessed, OLR-W\kern-0.25em A model achieves the lowest average rank of 1.25. This finding underscores the significance of further investigating the statistical implications of online regression models.

\begin{table*}[t]
\captionsetup{justification=centering}
\caption{ Convergence Analysis\\
{\scriptsize Summary of Performance Measures Using MSE on the First 10 Iterations \\  
\scriptsize $E$: Epochs, $K$: Mini-batch Size, $\mathbb{Z}^{+}$: Natural Number $\geq$ 1}}
\centering
\renewcommand{\theadalign}{cc} 
\renewcommand{\theadfont}{\small} 
\setlength{\tabcolsep}{6pt} 
\begin{adjustbox}{width=\textwidth} 
\begin{tabular}{|cc|*{16}{c|}}
\hline
\multirow{2}{*}{} & \multirow{2}{*}{} & \multicolumn{16}{c|}{\thead{Algorithms}} \\ \hline
{\thead{Datasets}} & & \multicolumn{2}{c|}{\rotatebox[origin=c]{90}{\thead{\textbf{SGD} \\ {\scriptsize $\eta=0.01$ [DS3, DS4 =0.001},\\ \scriptsize{ KCHSD =$1 \mathrm{e}{-4}$]}, \\ {\scriptsize E=N $\times$ $(\mathbb{Z}^{+}=2)$ }\\{\scriptsize [CCPP $\mathbb{Z}^{+}=5$]}}}} 
& \multicolumn{2}{c|}{\rotatebox[origin=c]{90}{\thead{\textbf{MBGD} \\ {\scriptsize $\eta=0.01$ [1KC= 0.1],}\\{\scriptsize$ E=\frac{N}{K} \times (\mathbb{Z}^{+} =5)$ [DS3, DS4 \vspace{1.5mm}}\\{\scriptsize  $\mathbb{Z}^{+}$=100, KCHSD $\mathbb{Z}^{+}$=20,}\\{\scriptsize 1KC, CCPP $\mathbb{Z}^{+}$=40]}}}} 
& \multicolumn{2}{c|}{\rotatebox[origin=c]{90}{\thead{\textbf{LMS} \\ {\scriptsize $\eta=0.01$[DS3, DS4=0.001},\\ \scriptsize{1KC=.1, KCHSD,CCPP=$1 \mathrm{e}{-4}$]}}}} 
& \multicolumn{2}{c|}{\rotatebox[origin=c]{90}{\thead{\textbf{ORR} \\ {\scriptsize $\eta=0.01$ [DS3, DS4, KCHSD,}\\ \scriptsize{=0.001], E=N $\times (\mathbb{Z}^{+}=2)$}\\{\scriptsize [1KC $\mathbb{Z}^{+}$=3, KCHSD, CCPP}\\ \scriptsize{ $\mathbb{Z}^{+}=5$], $\lambda=0.1$ [1KC=0.01,}\\{\scriptsize MCPD,KCHSD,CCPP=0.001]}}}} 
& \multicolumn{2}{c|}{\rotatebox[origin=c]{90}{\thead{\textbf{OLR} \\ {\scriptsize $\eta=0.01$ [DS3, DS4, KCHSD}\\{\scriptsize = 0.001], E=N $\times (\mathbb{Z}^{+}=2)$}\\{\scriptsize [1KC $\mathbb{Z}^{+}$=3, KCHSD, CCPP }\\{\scriptsize $\lambda=0.1$ [1KC, CCPP=0.01],}\\ \scriptsize{[ KCHSD=$1 \mathrm{e}{-4}$]}}}} 
& \multicolumn{2}{c|}{\rotatebox[origin=c]{90}{\thead{\textbf{RLS} \\ {\scriptsize $\lambda=.99$, $\delta=0.01$}\\{\scriptsize [1KC $\delta=0.1$]}}}}        
& \multicolumn{2}{c|}{\rotatebox[origin=c]{90}{\thead{\textbf{PA} \\ {\scriptsize \text{PA-}\uppercase\expandafter{\romannumeral3}}\\ {\scriptsize $C=0.1$, $\epsilon=0.1$},\\ \scriptsize{ [DS6, DS7, DS8}\\ \scriptsize{$C=0.01$, $\epsilon=0.01$]}}}} 
& \multicolumn{2}{c|}{\rotatebox[origin=c]{90}{\thead{\textbf{OLR-W\kern-0.25em A} \\ {\scriptsize $\alpha=.5$}\\{\scriptsize [KCHSD $\alpha=.9$}}}} \\ 
\hline
\thead{Name} & \thead{$K$} & \thead{MSE} & \thead{rank} & \thead{MSE} & \thead{rank} & \thead{MSE} & \thead{rank} & \thead{MSE} & \thead{rank} & \thead{MSE} & \thead{rank} & \thead{MSE} & \thead{rank} & \thead{MSE} & \thead{rank} & \thead{MSE} & \thead{rank} \\ \hline
\thead{DS1 \cite{OLR-WA_GIT}} & 10 & 1865.20581 & 4 & 5976.67276 & 8 & 1778.39859 & 3 & 1961.97400 & 6 & 1933.74953 & 5 & 3080.20632 & 7 & 541.27625 & 2 & 455.90216 & 1 \\ \hline
\thead{DS2 \cite{OLR-WA_GIT}} & 30 & 2430.97513 & 5 & 23760.09556 & 8 & 2390.53622 & 4 & 2825.94930 & 6 & 2122.09717 & 3 & 4715.74581 & 7 & 1564.91181 & 2 & 841.70463 & 1 \\ \hline
\thead{DS3 \cite{OLR-WA_GIT}} & 300 & 3282.66236 & 6 & 31822.14283 & 8 & 3275.53508 & 5 & 3887.32562 & 7 & 3218.66053 & 4 & 1593.63189 & 2 & 2349.97800 & 3 & 1165.10646 & 1 \\ \hline
\thead{DS4 \cite{OLR-WA_GIT}} & 1,000 & 5550.43115 & 5 & 31435.32987 & 8 & 5547.42274 & 4 & 5658.62652 & 6 & 5429.84169 & 3 & 12569.60925 & 7 & 5407.04210 & 2 & 3565.32212 & 1 \\ \hline
\thead{MCPD \cite{MCPD_DS}} & 25 & 0.33973 & 8 & 0.85781 & 8 & 0.31242 & 1 & 0.33924 & 3 & 0.33995 & 5 & 0.36409 & 7 & 0.36793 & 7 & 0.31687 & 2 \\ \hline
\thead{1KC \cite{1KC_DS}} & 39 & 0.00362 & 6 & 0.00709 & 8 & 0.00138 & 2 & 0.00358 & 4 & 0.00359 & 5 & 0.00151 & 3 & 0.00464 & 7 & 0.00097 & 1 \\ \hline
\thead{KCHSD \cite{KCHS_DS}} & 30 & 0.88983 & 7 & 0.69907 & 6 & 0.89519 & 8 & 0.59394 & 5 & 0.57859 & 4 & 0.49323 & 1 & 0.50757 & 3 & 0.50742 & 2 \\ \hline
\thead{CCPP \cite{CCPP_DS}} & 20 & 0.05089 & 4 & 0.18292 & 7 & 0.23790 & 8 & 0.05065 & 3 & 0.05058 & 2 & 0.09088 & 6 & 0.07098 & 5 & 0.00407 & 1 \\ \hline
\thead{Average} &  & 1,641.32 & 5.125 & 11,624.50& 7.625 & 1624.16744& 4.375 & 1,791.86& 5 & 1,588.17& 3.875 & 2,745.02& 4.875 & 1,233.02& 3.875 & 753.61  & 1.250 \\ \hline
\end{tabular}
\end{adjustbox}
\label{table:models_performance_first_10_minibatches}
\end{table*}

\subsection{Performance Analysis on Concept Drift Datasets.}

\begin{figure}[ht]
  \centering
  \includegraphics[width=0.60\textwidth, height=0.177\textheight]{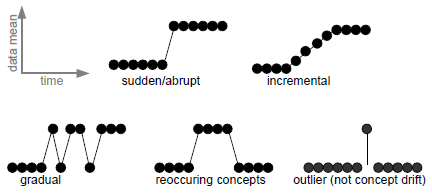}
  \caption{Concept Drift Types \cite{gama2014survey}}
  \label{fig:001_drift_types}
\end{figure}

In this section, four top-performing models are evaluated across diverse types of concept drift, as illustrated in Figure \ref{fig:001_drift_types}, using the concept drift datasets summarized in Table \ref{tab:drift-datasets-properties}. This experiment employs OLR-W\kern-0.25em A\kern-0.25em A, the specialized variant designed to handle concept drift through the automatic inference of the hyperparameter $\alpha$.

\begin{table}[!htbp]
\centering
\captionsetup{justification=centering}
\caption{Dataset Properties.\\ {\scriptsize
Drift Locations: Abrupt (A) marks sudden drift onset. Incremental (I) drift recurs every $K$ points. Gradual (G) shows drift as a sequence of concepts C$_1$ and C$_2$.
E.D.: Euclidean distance between coefficients in space. C-Start: Start Concept. C-End: End Concept. E.D. (Consec.): E.D. between each two consecutive concepts in sequence.}}
\renewcommand{\theadalign}{bc}
\renewcommand{\theadfont}{\bfseries}
\setlength{\tabcolsep}{3pt}
\renewcommand{\arraystretch}{1.2}
\scriptsize
\begin{tabularx}{\textwidth}{@{}
c  
c  
c  
c  
c  
c  
>{\centering\arraybackslash}m{1.7cm}  
>{\centering\arraybackslash}m{2.1cm}  
@{}}
\toprule
\thead{Dataset} & 
\thead{Drift\\Type} & 
\thead{Data-\\points} & 
\thead{Dimen-\\sions} & 
\thead{Noise} & 
\thead{Drift\\Locations} &
\thead{E.D.\\(C-Start\\to C-End)} &
\thead{E.D.\\(Consec.)} \\
\midrule

DS11\cite{OLR-WA_GIT} & A & 10k & 10 & 20 & 5k & 327.23 & \makecell[t]{327.23} \\[.5em]

DS12\cite{OLR-WA_GIT} & I & 10k & 10 & 20 & every 1k & 1205.10 & 
\makecell[t]{[14.8, 18.5, 23.9,\\ 31.8, 44.6, 66.9,\\ 111.5, 223.1, 669.5]} \\[1.2em]

DS13\cite{OLR-WA_GIT} & G & 10k & 10 & 20 & 
\makecell[t]{[2.5k-c1, 1k-c2,\\ 1k-c1, 2k-c2,\\ 1k-c1, 2.5k-c2]} & 
2049.14 & 
\makecell[t]{[1158.3, 1205.1,\\ 1885.4, 1885.4,\\ 2167.7]} \\[1em]

\bottomrule
\end{tabularx}
\label{tab:drift-datasets-properties}
\end{table}

\subsubsection{Performance Analysis on Abrupt Drift datasets}
\begin{figure}[H]
\centering
\includegraphics[width=1\textwidth]{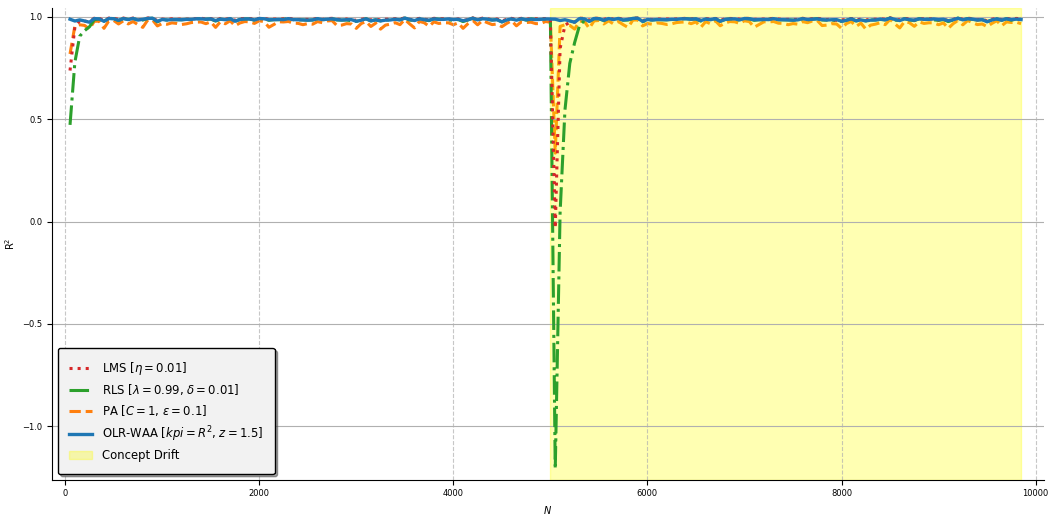}
\captionsetup{justification=centering}
\caption{Models Performance Analysis on Abrupt Drift Dataset DS11}
\label{fig:abrupt_drift_expr}
\end{figure}

\textit{Performance Evaluation:} Figure \ref{fig:abrupt_drift_expr} illustrates the performance of OLR-W\kern-0.25em A\kern-0.25em A, PA, RLS, and LMS on abrupt drift datasets, measured using the $R^2$ metric. The yellow region indicates the onset of a new drifted concept starting at data point 5000. OLR-W\kern-0.25em A\kern-0.25em A demonstrates superior robustness to abrupt drift, while PA, RLS, and LMS exhibit varying degrees of sensitivity, with RLS being the most affected.

It is important to note that RLS includes a hyperparameter $\lambda \in (0,1]$, known as the forgetting factor, which determines the relative weighting of recent versus past observations. A value of $\lambda = 1$ assigns equal importance to all data points, while $\lambda < 1$ applies exponential forgetting, emphasizing more recent data. In our online learning experimental setup, $\lambda$ is fixed at 0.99 by default to enable mild forgetting while avoiding any risk of data snooping. However, since RLS lacks a mechanism to dynamically adjust this parameter in response to drift, it may be suboptimal in non-stationary environments. This limitation is addressed in OLR-W\kern-0.25em A\kern-0.25em A, which incorporates an internal mechanism to automatically adjust hyperparameters upon each detected concept drift.

\subsubsection{Performance Analysis on Incremental Drift datasets}

\begin{figure}[H]
\centering
\includegraphics[width=1\textwidth]{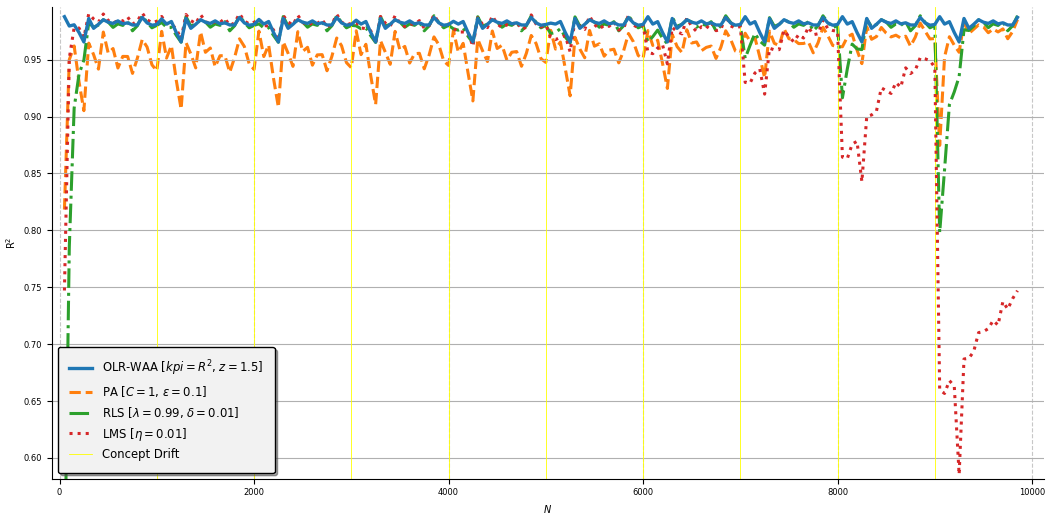}
\captionsetup{justification=centering}
\caption{OLR-W\kern-0.25em A\kern-0.25em A 
 Performance Analysis on Incremental Drift Dataset DS12}
\label{fig:incremental_drift_expr}
\end{figure}

\textit{Performance Evaluation:} Figure \ref{fig:incremental_drift_expr} illustrates the performance of the models under incremental drift, where the yellow lines indicate the onset of new concepts in an incremental manner. The models exhibit reduced sensitivity to incremental drift compared to abrupt drift, with OLR-W\kern-0.25em A\kern-0.25em A demonstrating the best overall performance. Additionally, LMS shows a significant degradation in performance beyond data point 8000, which may be attributed to the limitations of a fixed learning rate in such settings.

\subsubsection{Performance Analysis on Gradual Drift datasets} 

\begin{figure}[H]
\centering
\includegraphics[width=1\textwidth]{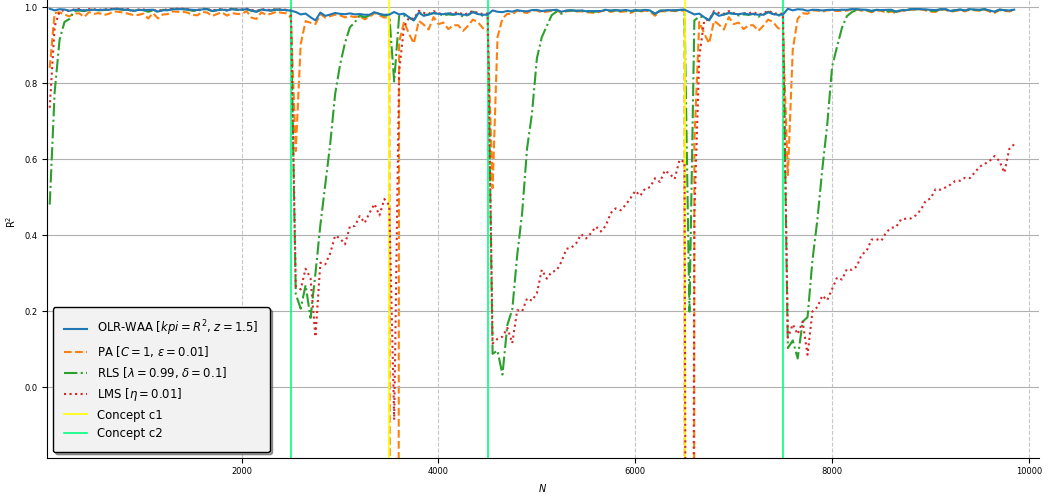}
\captionsetup{justification=centering}
\caption{OLR-W\kern-0.25em A\kern-0.25em A 
 Performance Analysis on Gradual Drift Dataset DS13}
\label{fig:gradual_drift_expr}
\end{figure}

\textit{Performance Evaluation:} Figure \ref{fig:gradual_drift_expr} illustrates the performance of the models on the gradual drift dataset DS13. The green line indicates the onset of concept 2, while the yellow line marks the reintroduction of concept 1. Overall, all models exhibit significant performance degradation, primarily due to fixed hyperparameters that lack adaptability in the presence of concept drift. Notably, OLR-W\kern-0.25em A\kern-0.25em A demonstrates superior resilience, emerging as a top-performing model in handling gradual concept drift.

\section{Conclusion and Future Work}
In streaming environments, batch learning is often unsuitable due to its reliance on full data access, ample training time, and the assumption of a stationary \textit{i.i.d.} distribution. For instance, in the stock market, models trained on outdated historical data may fail to capture rapidly evolving patterns. Online learning addresses these challenges by enabling real-time updates from sequential data and handling distribution shifts. However, it faces limitations such as difficulty adapting to concept drift and the impracticality of manually tuning hyperparameters during continuous change.

This paper presents OLR-W\kern-0.25em A\kern-0.25em A: Adaptive and Drift-Resilient Online Regression with Dynamic Weighted Averaging, a hyperparameter-free model designed for robust, real-time regression in evolving data streams. OLR-W\kern-0.25em A\kern-0.25em A incrementally updates its base model by integrating streaming data using an exponentially weighted moving average, facilitating smooth adaptation to evolving patterns. To address concept drift, OLR-W\kern-0.25em A\kern-0.25em A introduces a novel optimization mechanism that proactively detects and responds to drift using real-time, in-memory processing. This mechanism incorporates holistic drift handling by combining drift detection, severity quantification, adaptive hyperparameter adjustment. It is designed for large-scale, high-dimensional datasets by relying on performance-driven KPIs without making distributional assumptions. Additionally, the incorporation of dynamic thresholding that does not assume any prior data distribution enhances the effectiveness of OLR-W\kern-0.25em A\kern-0.25em A in adapting to varying data patterns, managing high-dimensional inputs, and supporting diverse performance KPIs.

We conducted four types of experiments: stationary, adversarial, convergence, and concept drift, across a diverse set of datasets. The empirical evaluation establishes OLR-W\kern-0.25em A  and its variant OLR-W\kern-0.25em A\kern-0.25em A as a top-tier model for handling a diverse range of scenarios. It ranked consistently top in stationary tasks and excelled in both time-based and uniquely in confidence-based adversarial scenarios. In convergence, it was the only model to achieve low cost and continuously sustain that diminished cost throughout. In contrast to other online models, which exhibit fluctuating performance on different concept drift datasets, OLR-W\kern-0.25em A\kern-0.25em A consistently demonstrated superior robustness across all types of concept drift.


Future efforts will concentrate on expanding the OLR-W\kern-0.25em A\kern-0.25em A approach to a broader spectrum of machine learning tasks, aiming to boost its versatility and applicability across various domains. Concurrently, we plan to implement OLR-W\kern-0.25em A\kern-0.25em A as an online web service for drift-aware, real-time classification, thereby embracing the principles of Machine Learning as a Service (MLaaS).


\bibliography{sn-bibliography}


\end{document}